\documentclass[10pt,twocolumn,letterpaper]{article}
\usepackage[british,english,american]{babel}
\usepackage[pagenumbers]{cvpr}              % To produce the CAMERA-READY version
\usepackage{multirow}
\usepackage[table,xcdraw]{xcolor}
\usepackage{float}

\renewcommand{\paragraph}[1]{\vspace{1.25mm}\noindent\textbf{#1}}
\newcommand{\tablestyle}[2]{\setlength{\tabcolsep}{#1}\renewcommand{\arraystretch}{#2}\centering\footnotesize}
\newcommand{\ali}[1]{}

\newcommand{\app}{\raise.17ex\hbox{$\scriptstyle\sim$}}

\newlength\savewidth

\newcolumntype{x}[1]{>{\centering\arraybackslash}p{#1pt}}
\newcolumntype{y}[1]{>{\raggedright\arraybackslash}p{#1pt}}
\newcolumntype{z}[1]{>{\raggedleft\arraybackslash}p{#1pt}}

\definecolor{mygreen}{HTML}{B3D344}

\newcommand\deemph{\color[gray]{0.6}}

\newcommand\textdes[1]{{\footnotesize\textsf{{#1}}}}

\definecolor{cvprblue}{rgb}{0.21,0.49,0.74}
\usepackage[breaklinks,colorlinks,allcolors=cvprblue]{hyperref}

\title{ARC Is a Vision Problem! \vspace{-.4em}}

\author{Keya Hu \quad Ali Cy \quad Linlu Qiu \quad Xiaoman Delores Ding \\ Runqian Wang \quad Yeyin Eva Zhu \quad Jacob Andreas \quad Kaiming He \\[.3em]
MIT }

\begin{document}
\maketitle
\vspace{-1em}
\begin{abstract}
The Abstraction and Reasoning Corpus (ARC) is designed to promote research on abstract reasoning, a fundamental aspect of human intelligence.
Common approaches to ARC treat it as a language-oriented problem, addressed by large language models (LLMs) or recurrent reasoning models.
However, although the puzzle-like tasks in ARC are inherently visual, existing research has rarely approached the problem from a vision-centric perspective.
In this work, we formulate ARC within a vision paradigm, framing it as an image-to-image translation problem.
To incorporate visual priors, we represent the inputs on a “canvas” that can be processed like natural images.
It is then natural for us to apply standard vision architectures, such as a vanilla Vision Transformer (ViT), to perform image-to-image mapping.
Our model is trained from scratch solely on ARC data and generalizes to unseen tasks through test-time training.
Our framework, termed \mbox{Vision ARC} (\textbf{VARC}), achieves 60.4\% accuracy on the \mbox{ARC-1} benchmark, substantially outperforming existing methods that are also trained from scratch.
Our results are competitive with those of leading LLMs and close the gap to average human performance.\footnote{Project webpage: \url{https://github.com/lillian039/VARC}.
}
\end{abstract}
\vspace{-1em}    
\section{Introduction}
\label{sec:intro}

\begin{figure}
\vspace{-1em}
\centering
\includegraphics[width=0.95\linewidth]{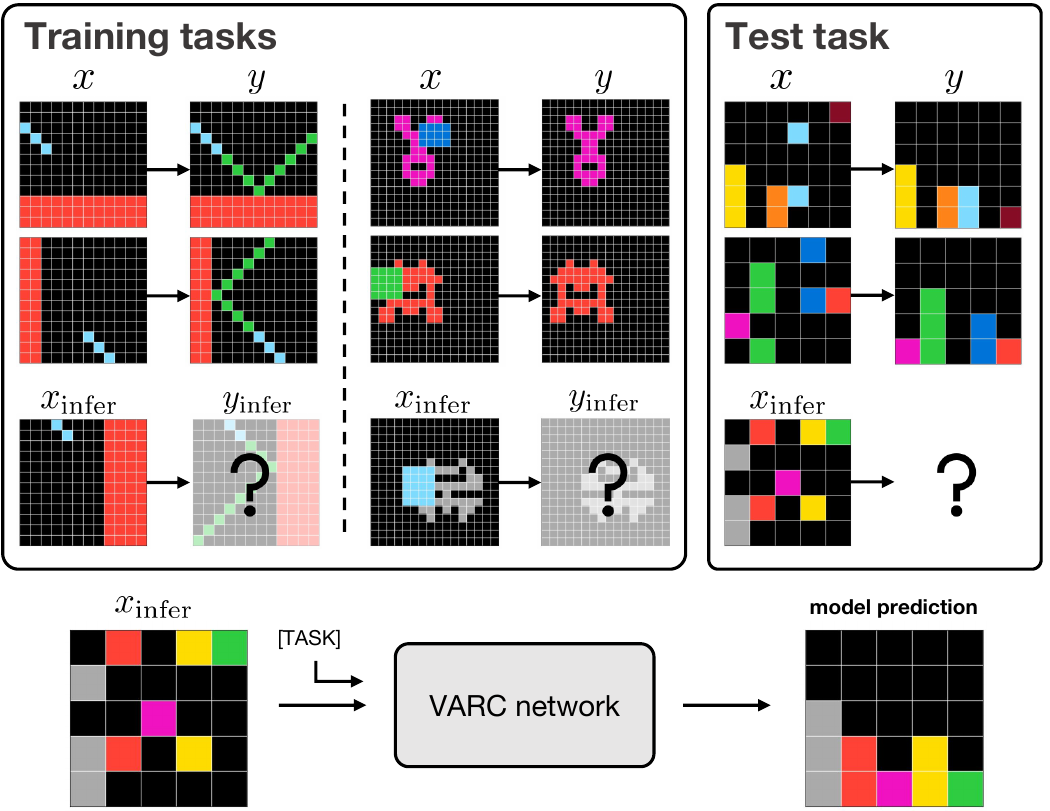}
\vspace{-0.5em}
\caption{The \textbf{ARC benchmark} (top) consists of a collection of many different tasks, where each task has a few (\eg, 2-4) examples.
We propose the {Vision ARC} (\textbf{VARC}) framework, which addresses the ARC problem as an image-to-image translation problem, from a computer vision perspective (bottom). In this illustration, the underlying concepts of the three tasks can be roughly described by humans as: {``\textdes{reflection}''} (left), {``\textdes{symmetry}''} (middle), and {``\textdes{gravity}''} (right). These concepts are closely related to the visual and physical world.
}
\label{fig:arcexamples}
\vspace{-.5em}
\end{figure}

Learning and abstracting concepts from a small number of demonstrations is a key feature of intelligence.
The Abstraction and Reasoning Corpus (ARC) benchmark \cite{chollet2019measureintelligence} was designed to incentivize machine learning research aimed at improving these capabilities.
ARC consists of a collection of puzzle-like tasks (\cref{fig:arcexamples}, top), each containing only a few examples governed by a unique underlying transformation rule. The model is expected to make predictions on each \mbox{\textit{unseen}} task given a few examples.
While humans are capable of solving various ARC tasks \cite{johnson2021fastflexiblehumanprogram, legris2024harcrobustestimatehuman, LeGris2025HARC_SciData}, the benchmark remains highly challenging for today's leading machine learning systems~\cite{pfister2025understanding, moskvichev2023conceptarc}.

The ARC problem has attracted significant attention, and substantial progress has been made in recent years~\cite{chollet2025arcprize2024technical}.
Among a wide variety of methods, those based on large language models (LLMs) have proven highly competitive.
These methods generally convert ARC inputs into sequences of text tokens for language modeling.
Representative methods may involve inductive reasoning \cite{wang2024hypothesis,berman2024record536, tang2024code,berman2024arcagi},  transductive reasoning \cite{akyurek2025surprising, franzen2025productexpertsllmsboosting, puget2024arc}, or a combination of both \cite{li2025combining, berman2025highestscore, macfarlane2025searchinglatentprogramspaces}. The LLMs are pre-trained on internet-scale data, from which they learn transferable common sense. 

Most recently, research on \textit{recurrent} models \cite{wang2025hierarchicalreasoningmodel,jolicoeurmartineau2025morerecursivereasoningtiny} has achieved impressive results on ARC without relying on internet-scale data. These models are trained from scratch on ARC data only and perform inference through recurrent, iterative reasoning. 
Although they do not rely on large-scale language pre-training, these recurrent models draw strong inspiration from the success of language modeling.

% ####################################################################################################
\begin{figure}[t]
\vspace{-1em}
\centering
\includegraphics[width=1.0\linewidth,clip,trim={0.15cm 21cm 6cm 0.15cm}]{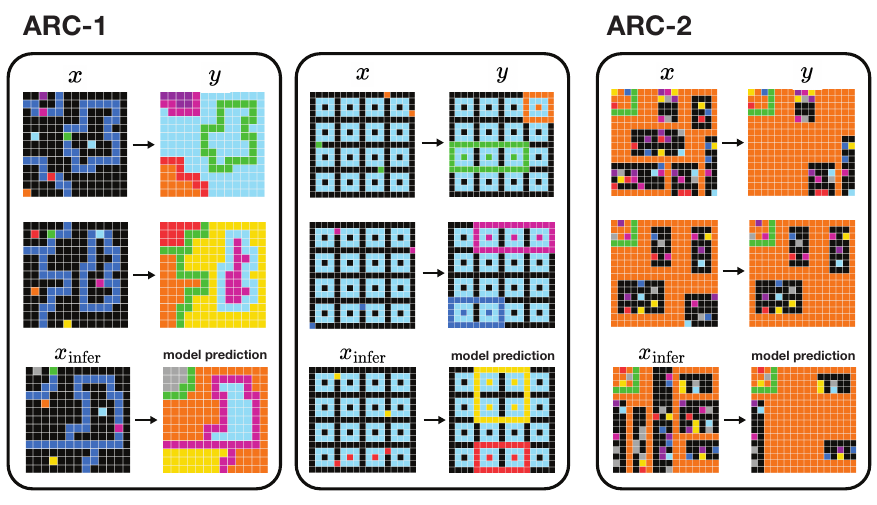}
% \vspace{-1em}
\caption{
\textbf{Examples of unseen tasks solved by VARC.}
Each panel shows an unseen test task, with demonstrations on the top and the model’s prediction on the bottom. VARC correctly solves these challenging tasks.
}
\label{fig:solved_examples_main}
\vspace{-.5em}
\end{figure}
% ####################################################################################################

Interestingly, although the ARC puzzles are typically presented visually, existing research has rarely framed ARC as a vision-centric problem.
In fact, many concepts in ARC are inherently \textit{visual} and \textit{physical}: \eg, reflection, symmetry, and gravity, as shown in \cref{fig:arcexamples}. Humans can solve these tasks not merely from the demonstrations, but by reasoning through analogy to their common sense obtained from external experience.
Such common sense can be acquired through observing the world, particularly, the \textit{visual} world.

Motivated by its visual nature, we approach ARC from a \textit{vision-centric} perspective.
We frame each puzzle as an image-to-image translation problem.
Abstraction and inference can arise directly from visual learning, without explicit linguistic intermediates.
This perspective connects ARC to classical image-to-image problems, ranging from low-level image processing (\eg, \cite{dong2015image,pathak2016context}) to high-level image understanding (\eg, \cite{long2015fully,ronneberger2015unetconvolutionalnetworksbiomedical}).
With this connection, we can apply standard vision models (\eg, Vision Transformers~\cite{dosovitskiy2021an} or convolutional networks~\cite{6795724}) to tackle the ARC problem.

We demonstrate that incorporating \textit{visual priors} is crucial.
These priors include 2D spatial locality, translation invariance, and scale invariance.
To facilitate learning these priors, we represent the inputs on a ``canvas'' with flexible geometric transformations, allowing the inputs to be processed as if they were natural images.
A patch on the canvas can consist of exponentially many color combinations, which helps reduce overfitting and encourages the model to learn spatial priors rather than merely memorize.

With the vision-centric formulation, we train our model \textit{from scratch} using ARC-only data. 
At inference time, when presented with a new, unseen task, we perform test-time training \cite{bottou1992locallearning,joachims1999transductive,Sun2020TTT,akyurek2025surprising,wang2025hierarchicalreasoningmodel,jolicoeurmartineau2025morerecursivereasoningtiny} to adapt the model to the task, enabling it to generalize from only a few examples.

Our framework, termed {Vision ARC} (\textbf{VARC}), shows strong performance on the ARC benchmarks (\eg, \cref{fig:solved_examples_main}).
VARC achieves 54.5\% accuracy on the ARC-1 benchmark, using a small model with only 18 million parameters. This result substantially surpasses the best recurrent methods \cite{wang2025hierarchicalreasoningmodel,jolicoeurmartineau2025morerecursivereasoningtiny} that are also trained from scratch on ARC. 
It is also competitive with many popular LLM-based methods.
Combining VARC models through ensembling \cite{krizhevsky2012imagenet} further improves accuracy to 60.4\%, matching the reported average human performance \cite{legris2024harcrobustestimatehuman} on the ARC-1 dataset.

We hope our research will shed light on the ARC problem, and more broadly, on the field of abstract reasoning. On the one hand, the design of the ARC benchmark is based on human observations and induced rules abstracted from the visual and physical world. It is natural to explore vision-driven approaches for ARC. On the other hand, human reasoning is not confined to language or vision in isolation, but instead should integrate information across modalities. With our complementary vision-based perspective, we hope the scope of abstract reasoning will be further broadened. We invite the vision community to study the ARC problem and to advance research on abstract reasoning.

\section{Related Work}

\vspace{-.5em}
\paragraph{Visual reasoning.}
Visual reasoning is a long-standing research problem.
It involves not only perceiving scenes and objects, but also inferring and abstracting the relations and transformations among them.
The advancement of machine learning methods has led to the development of a variety of challenging protocols, such as VQA~\cite{antol2015vqa, zhang2016balancing, goyal2017making}, CLEVR~\cite{johnson2017clevr}, and Winoground~\cite{thrush2022winoground}.

The visual reasoning methods developed under these protocols typically consist of a visual perception module and a language-like recurrent module, \eg, within the neuro-symbolic framework~\cite{andreas2016neural, hu2017learning, andreas2016learning, mao2019neuro}.
These methods have evolved into modern vision-language models (VLMs, \eg, \cite{alayrac2022flamingo, li2022blip,liu2023visual}), in which images are converted into tokens and processed jointly with text.

Unlike ARC, classical visual reasoning protocols generally involve a training set and a test set, both of which can be viewed as instances of \textit{the same} task. In contrast, ARC consists of a large collection of distinct tasks, each defined by only a few examples.

\paragraph{Approaches to ARC.} Owing to the ``\textit{few-shot, many-task}'' nature of ARC, LLMs have been regarded as a natural solution.
A new task can be converted into a sequence of tokens, treated as a prompt, and processed by LLMs via in-context few-shot learning~\cite{wei2022chain, brown2020language}. We refer the reader to \cite{chollet2025arcprize2024technical} for a comprehensive survey.

Recently, \textit{recurrent} models \cite{wang2025hierarchicalreasoningmodel,jolicoeurmartineau2025morerecursivereasoningtiny} have been proven effective for ARC, without the requirement of internet-scale pre-training.
These models aim to mimic the hierarchical and multi-timescale processing of the human brain \cite{wang2025hierarchicalreasoningmodel} for reasoning.
At inference time, these methods adopt test-time training \cite{bottou1992locallearning,joachims1999transductive,Sun2020TTT} on the few demonstration examples. 

Related to our work, the ViT-ARC method \cite{li2025tacklingabstractionreasoningcorpus} attempts to address the ARC problem using vision models.
However, this method has only shown the ability to fit individual tasks in the training set; it is unable to generalize or solve any unseen test task.
As such, this method has not been able to satisfy the ARC protocol, whose essence lies precisely in few-shot, \textit{cross-task} generalization.
Unlike \cite{li2025tacklingabstractionreasoningcorpus}, our framework is designed to address the “few-shot, many-task” nature of ARC.

\section{ARC as a Vision Problem}

\begin{figure}[t]
\centering
\includegraphics[width=1.0\linewidth]{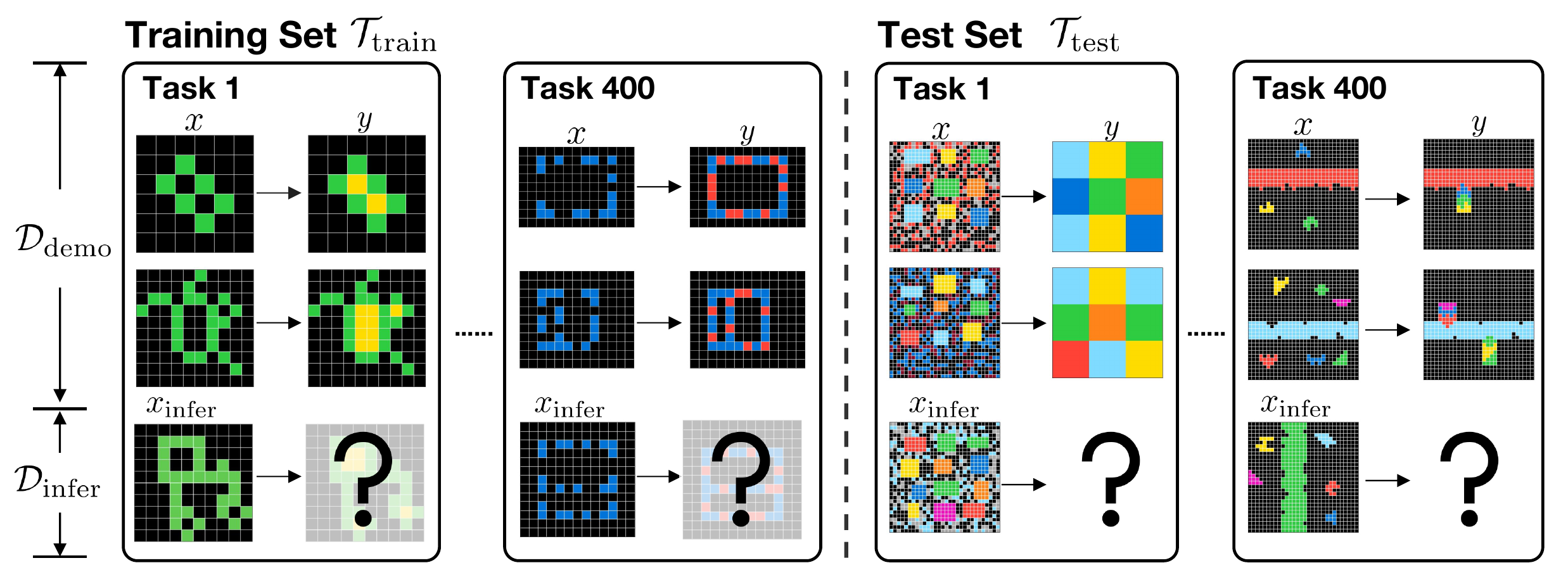}
\vspace{-1.5em}
\caption{\textbf{The ARC problem definition}. ARC is a collection of many different tasks.
For each task, a few (\eg, 2-4) demonstration pairs $(x, y)$ are given, and the model is required to infer the output from $x_\textrm{infer}$.
The training set $\mathcal{T}_\textrm{train}$ is a collection of 400 tasks, which can be used for model training. The test set $\mathcal{T}_\textrm{test}$ contains 400 new tasks: the demo pairs of a new task are given only at inference time, based on which the model performs inference on $x_\textrm{infer}$.
}
\label{fig:train_protocal}
\vspace{-0.6em}
\end{figure}

\subsection{ARC Problem Definition}

The ARC benchmark consists of several hundred very few-shot (\eg, 2 to 4-shot) reasoning tasks. 
Each task, denoted by $T$, involves a unique underlying transformation rule, mapping from an input $x$ to an output $y$.
Here, $x$ and $y$ are both 2D grids with maximum size $30{\times}30$, in which each location has one of $C$ different color indexes (\eg, $C{=}10$). The ARC problem definition is illustrated in \cref{fig:train_protocal}, which we discuss next.

\paragraph{A task.} A ``task'' is the basic unit in ARC.
Each task includes a few \textit{demonstration} examples. For a demonstration pair $(x, y)$, both $x$ and $y$ are known to the model. We denote the demonstration set of task $T$ as: $\mathcal{D}_\textrm{demo}^T {=} \big\{(x_i, y_i)\big\}_{i=1}^m$, where $m$ is the number of pairs (\eg, $m$ is 2 to 4).
Each task $T$ also contains a few \textit{inference} examples, denoted as: $\mathcal{D}_\textrm{infer}^T {=} \big\{(x_i, y_i)\big\}_{i=1}^n$ ($n$ is 1 or 2). At inference time, only the demo pairs $\mathcal{D}_\textrm{demo}^T$ and one input $x_\textrm{infer}\in \mathcal{D}_\textrm{infer}^T$ are given, and the model is required to infer the desired output $y_\textrm{infer}$.

\paragraph{Training set.} The training set consists of multiple tasks used to train the model \textit{offline} (\ie, before a new task is given). We denote the training set as: $\mathcal{T}_\textrm{train}{=}\{ T_i \}_{i=1}^k$, where $k$ is the number of tasks (400 in ARC-1). 
Following standard machine learning protocols, samples in $\mathcal{D}_\textrm{demo}^T$ for any $T\in\mathcal{T}_\textrm{train}$ can be used for training.
The ``inference'' samples in the training set, that is, $\mathcal{D}_\textrm{infer}^T$ for any task $T \in \mathcal{T}_\textrm{train}$, are used for validating the training process only.

\paragraph{Test set.} The test set is a collection of \textit{new} tasks, which are not seen during offline training. We denote the test set as: $\mathcal{T}_\textrm{test}{=}\{ T_i \}_{i=1}^l$, with $l$ different test tasks.
Note that any test task is a ``complete'' and new task: that is, for any $T\in\mathcal{T}_\textrm{test}$, there also exists a demo set $\mathcal{D}_\textrm{demo}^T$, and the pairs $(x, y)$ in $\mathcal{D}_\textrm{demo}^T$ are given to the model at inference time.
The model should make use of $\mathcal{D}_\textrm{demo}^T$ to infer the output of the given
$x_\textrm{infer}$ for this new task.

The presence of new $(x, y)$ pairs in $\mathcal{D}_\textrm{demo}^T$ at inference time allows to perform test-time training \cite{Sun2020TTT,akyurek2025surprising,bottou1992locallearning,joachims1999transductive}, which we adopt and will discuss.

\subsection{Image-to-Image Translation}

With these definitions, we formulate reasoning on each task as an image-to-image translation problem. We frame the problem as per-pixel classification, analogous to the semantic segmentation problem \cite{long2015fully}.

Formally, we learn a neural network $f_\theta$ parameterized by $\theta$. The network $f_\theta$ takes an image $x_i$ as input, conditioned on a task token associated with the task $T$. The task token is represented as a learnable embedding dependent on $T$. The output of $f_\theta$ is a grid where each position represents a categorical distribution.
The overall objective function is simply the per-pixel cross-entropy loss \cite{long2015fully}:
\begin{equation}
    \mathcal{L}(\theta) = \mathbb{E}_{T,i} \big[ \mathcal{D}(y_i,\,f_{\theta}( x_i \mid  T)) \big].
\label{eq:loss}
\end{equation}
Here, $\mathcal{D}$ denotes the per-pixel cross-entropy loss between the ground-truth $y_i$ and the network output.

\subsection{Visual Modeling}

Previous methods on ARC generally operate in the space of discrete-valued tokens, motivated by the design of language models. In our formulation of image-to-image translation, we explore \textit{native} designs developed for vision.

\paragraph{Canvas.} While it is straightforward to view the raw $H{\times}W$ grid as an $H{\times}W$ image, we propose more flexible transformations to represent it in a manner similar to natural images.

We define the concept of a ``\textit{canvas}''. A canvas has a predefined and sufficiently large size, \eg, $64{\times}64$. The raw input is transformed and placed onto this canvas. This formulation naturally accommodates translation and scale augmentations, which are common strategies for introducing translation and scale invariance in vision, discussed next.
We set the background of the canvas to an additional background color, \ie, the $(C{+}1)$-th color.

\begin{figure}[t]
    \centering
    \includegraphics[width=0.9\linewidth]{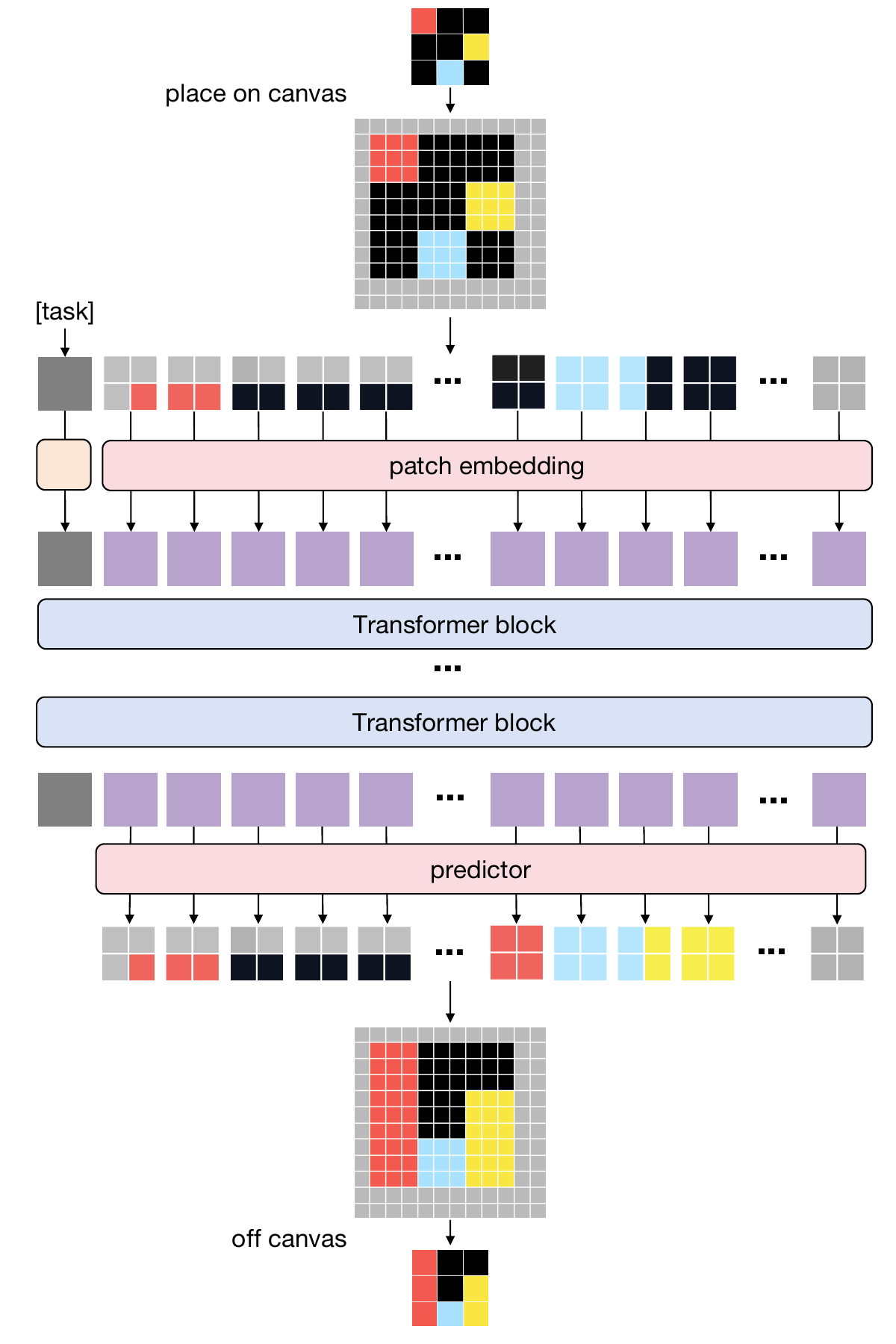}
    \vspace{-0.8em}
    \caption{
    \textbf{The ViT architecture in VARC}. The input is randomly placed on a canvas, which is then treated as a natural image and processed by a standard ViT, conditioned on the task token.}
    \label{fig:model_structure}
    \vspace{-0.8em}
\end{figure}

When applying a ViT model (discussed next), if we na\"ively treat each raw pixel as a token, there would be only $C$ distinct tokens.
In contrast, our canvas formulation supports a much larger set of local, patch-level configurations.
For example, with a patch size of $2{\times}2$ (see \cref{fig:model_structure}), a single patch can contain multiple colors and, in principle, has an exponentially large cardinality, $O(C^{2{\times}2})$.
This formulation is important for improving generalization performance.

% ####################################################################################################
\begin{figure}[t]
    \centering
    \includegraphics[width=0.9\linewidth]{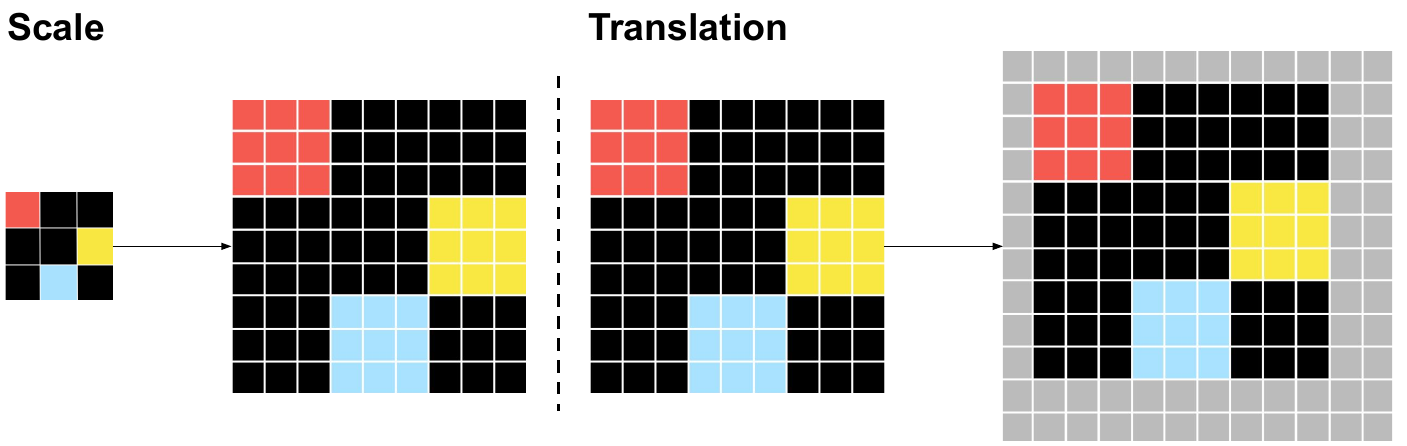}
    \vspace{-0.5em}
    \caption{The raw input undergoes random scale and translation transformations and is placed on the ``\textbf{canvas}'' (denoted in gray).}
    \label{fig:trans_res_aug}
    \vspace{-.5em}
\end{figure}
% ####################################################################################################

\paragraph{Translation and scale invariance.} The ``canvas'' concept enables us to flexibly apply translation and scale augmentations, which are critical in standard vision models. Theses data augmentations encourage the model to learn underlying mappings invariant to geometric transformations grounded in the visual world. Formally, we perform:

\begin{itemize}[leftmargin=1em, rightmargin=1em, itemsep=2pt, topsep=2pt]
  \item \textit{Scale augmentation}:
  Given a raw input, we randomly resize it by an integer scaling ratio $s$, duplicating each raw pixel into $s{\times}s$ (see \cref{fig:trans_res_aug}, left). This is analogous to nearest-neighbor interpolation in natural images. However, note that ``colors'' in ARC do not correspond to real-world colors, so it is not meaningful to perform other interpolations (such as bilinear). 
  \item \textit{Translation augmentation}: given the scaled grid, we randomly place it on the fixed-size canvas. We ensure all pixels are visibile. See \cref{fig:trans_res_aug} (right).  
\end{itemize}

\noindent
We empirically show that these visual priors are important for generalization to unseen tasks.

\paragraph{Vision Transformer.} Given a canvas with an input randomly placed, we perform image-to-image translation by a standard vision model. By default, we use a ViT \cite{dosovitskiy2021an}.

The principle of ViT is Transformer on patches. Formally, the input canvas is divided into non-overlapping patches (\eg, 2$\times$2), projected by a linear embedding,
added with positional embedding \cite{vaswani2023attentionneed}, and processed by a stack of Transformer blocks \cite{vaswani2023attentionneed}. The model has a linear projection layer as the output, which performs per-pixel classification for each patch.
Note that unlike natural images where each raw pixel has continuous values, in our case, the raw pixels have discrete values. Therefore, before patchification, we first map each pixel's discrete index into a learnable continuous-valued embedding.

% ####################################################################################################
% ####################################################################################################

Conceptually, patchification can be viewed as a special form of convolution. Like convolution, it incorporates several critical inductive biases in vision: most notably, {locality} (\ie, grouping nearby pixels) and {translation invariance} (\ie, weight sharing across locations).

\paragraph{2D positional embedding.} Unlike language data, which is generally modeled as 1D sequences, images are inherently 2D. This 2D structure can be lost if we na\"ively treat the embedded patches as a 1D sequence. We empirically show that explicitly modeling positions in 2D is essential.

Formally, we adopt \textit{separable} 2D positional embeddings, following \cite{chen2021empirical}: with $D$ channels for positional embeddings, we use the first half of the channels to embed the horizontal coordinate and the second half to embed the vertical coordinate.
This can be applied both to additive positional embeddings for encoding absolute positions and to the encoding of relative positions (\eg, RoPE \cite{su2023roformerenhancedtransformerrotary}).

\paragraph{Alternative: convolutional networks.}
Beyond ViT, we also study the more classical vision-based architecture, \ie, convolutional neural networks \cite{6795724}. Specifically, we adopt the U-Net model \cite{ronneberger2015unetconvolutionalnetworksbiomedical}, a hierarchical convolutional network. The original U-Net was proposed precisely for the image-to-image translation problem of segmentation \cite{ronneberger2015unetconvolutionalnetworksbiomedical}, making it a natural candidate for the problem we consider.

\subsection{Two-stage Training}

We adopt a two-stage training paradigm to learn the parameters of the neural network.

\paragraph{Offline training.} This stage is applied on the entire training set $\mathcal{T}_\textrm{train}$. It is on all demos $\mathcal{D}_\textrm{demo}^T$ for any $T\in\mathcal{T}_\textrm{train}$. We train one model $f_\theta$ jointly for all $k$ training tasks (\eg, $k{=}400$), based on the loss in \cref{eq:loss}. All tasks share the same parameters, only except that each task has its own task-conditional token.
We do not use the inference set $\mathcal{D}_\textrm{infer}^T$ from the training tasks (\ie, $T \in \mathcal{T}_\textrm{train}$) to train the model. These sets are used only for validation purposes.

% ######################################################################
\begin{figure}[t]
\centering
\vspace{-1em}
\includegraphics[width=1.0\linewidth]{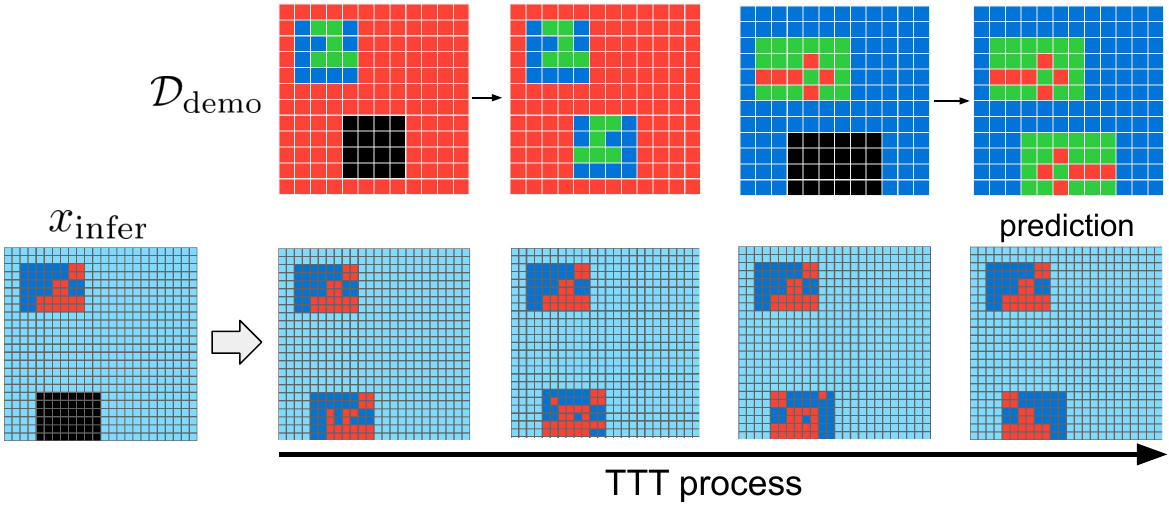}
\vspace{-1.8em}
\caption{
\textbf{Effect of test-time training.} (Top): Demonstration examples for the current task. (Bottom left): An inference example $x_\textrm{infer}$. (Bottom right): During test-time training, the prediction from $x_\textrm{infer}$ becomes progressively more accurate, with the model finally generating the correct prediction. 
}
\label{fig:ttt_example}
\vspace{-.5em}
\end{figure}
% ######################################################################

\paragraph{Test-time training (TTT).} Given a single new, unseen task $T\,\in\, \mathcal{T}_\textrm{test}$ from the \textit{test} set, we perform inference by test-time training.
At inference time, we are given $\mathcal{D}_\textrm{demo}^T {=} \big\{(x_i, y_i)\big\}_{i=1}^m$ with both input and output accessible; the model is required to make prediction for a given $x_\textrm{infer}$ in this new task $T$. The test-time training followed by inference can be viewed abstractly as a function \mbox{$\mathcal{F}(x_\textrm{infer} \mid \mathcal{D}_\textrm{demo}^T) \mapsto y_\textrm{infer}$}.

We perform test-time training for each new task $T$ \textit{independently}.
It has a new task token whose parameters are randomly initialized.
As there are very few demo pairs in $\mathcal{D}_\textrm{demo}^T$ (\eg, 2 to 4), we also perform data augmentation. We elaborate on the details in the next section and in appendix.

In summary, at inference time, the model is initialized from offline training, fine-tuned with test-time training only for the single new task $T$, and then performs inference on $x_\textrm{infer}$. As the new demo pairs in $\mathcal{D}_\textrm{demo}^T$ are very few, even with data augmentation, this test-time training process remains reasonably fast (\eg, 70 seconds per task on a single GPU).
\cref{fig:ttt_example} visualizes the effect of test-time training.

\subsection{Inference}
\label{subsec:inference}

After test-time training, we apply $f_\theta$ to $x_\textrm{infer}$ to obtain the final prediction. This process is analogous to the classical recognition problems \cite{krizhevsky2012imagenet,long2015fully}. Accordingly, we adopt post-processing strategies inspired by recognition methods.

\paragraph{Single-view inference.} Given $x_\textrm{infer}$ and a single ``view'' (\ie, with a given scale and translation), we place $x_\textrm{infer}$ on the canvas and apply $f_\theta$ to predict the output. Since one output location in the raw grid may be predicted by multiple pixels on the canvas (\eg, due to rescaling; see \cref{fig:model_structure}), we aggregate all predictions (from softmax outputs) at this location by average pooling. 

\paragraph{Multi-view inference.} It was a common practice to consolidate the predictions from multiple views (\eg, see AlexNet \cite{krizhevsky2012imagenet}).
Analogously, we adopt multi-view inference to improve accuracy, where the views are sampled with different augmentations. As the multi-view inference cost is negligible compared with test-time training cost, it is virtually nearly \textit{free} to use many views. We use 510 random views (details are in appendix). Predictions from different views are consolidated by \textit{majority voting} \cite{akyurek2025surprising}.\footnotemark{}
  
\footnotetext{In majority voting, two output grids are considered ``\textit{consistent}'' only when they are identical across the entire grid. The winner is the grid that is ``consistent'' with the largest number of other output grids.}

\paragraph{Pass@2 accuracy.}
The ARC benchmark by default adopts the pass@2 accuracy metric: \ie, two different solutions can be produced for evaluation, and a task is considered correct if one is correct. To support this metric, we adopt majority voting in multi-view inference and retain the top-2 most populated output solutions.

\section{Implementation Details}
\label{sec:impl}

We describe the major implementation choices in this section. The configuration details can be found in appendix.

\paragraph{Canvas.} In our best-performing model, the canvas size is $64{\times}64$. In the case of ViT, the patch size is $2{\times}2$, resulting in a sequence length of $32^2$.
For scale augmentation, an integer scaling ratio is randomly sampled, such that the scaled grid is no larger than the canvas size.
For translation augmentation, the upper-left corner is randomly sampled under the constraint that the placed image is fully visible.

\paragraph{Offline training.} We use the standard ARC-1 training set $\mathcal{T}_\textrm{train}$ for training: it has 400 tasks with 2-4 demo pairs each.
Following common practice on ARC, we also expand our training set with the RE-ARC set~\cite{hodel2024addressingabstractionreasoningcorpus}, from which we sample 1,000 additional demo pairs per task. Put together, our full training set has about 400k sample pairs. We apply translation and scale augmentation in offline training.

\paragraph{Test-time training.} Given an unseen task $T\in\mathcal{T}_\textrm{test}$, we have 2-4 sample pairs in $\mathcal{D}_\textrm{demo}^T$. To make test-time training more feasible, we also augment the single task $T$ into multiple \textit{auxiliary} tasks. We do this by using standard augmentation from existing ARC methods: flip, rotation (by 90$^\circ$, 180$^\circ$, or 270$^\circ$), and color permutation.
We treat each of these test-time training augmentations as an auxiliary task, each assigned a task embedding.
We also apply translation and scale augmentation in test-time training, but we do not view them as a new auxiliary task (under the assumption that all auxiliary tasks are translation and scale invariant).

\section{Experimental Results}

Our experiments are primarily conducted on the benchmark of \mbox{ARC-1}~\cite{chollet2019measureintelligence}. We report the \textbf{pass@2} accuracy (referred to simply as ``accuracy'' hereafter) in percentage (\%). To support pass@2 evaluation, we adopt multi-view inference. We also report final results on ARC-2~\cite{chollet2025arcagi2newchallengefrontier}.

We evaluate our model on the ARC-1 \textit{evaluation} set (\ie, $\mathcal{T}_\textrm{eval}$). This set is conceptually a test set (see \cref{fig:train_protocal}), but with ground truth available only for computing accuracy.

\begin{figure}[t!]
\centering
\includegraphics[width=1.0\linewidth, clip, trim=1em 2em 1em 0]{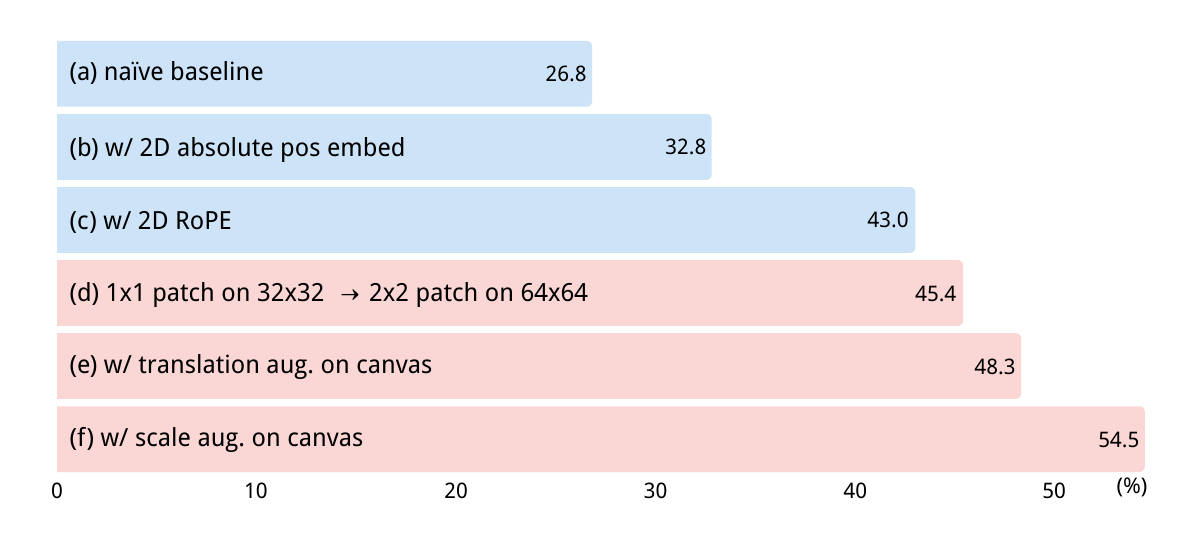}
% \vspace{-1.5em}
\caption{\textbf{Effects of visual priors in VARC}. Accuracy is reported on the ARC-1 evaluation set. The model used is ViT-18M. Entries (a-c) use a patch size of $1{\times}1$ on a $32{\times}32$ canvas, whereas entries (d-f) use a patch size of $2{\times}2$ on a $64{\times}64$ canvas. Each entry modifies the one above it. We start from a na\"ive baseline with components (b-f) removed. These vision priors cumulatively yield \textbf{27.7} improvement (a$\to$f), in which the canvas-based designs (c$\to$f) contribute an \textbf{11.5} gain.
}
\label{fig:visual_prior}
% \vspace{-1em}
\end{figure}

\subsection{Visual Priors}

\cref{fig:visual_prior} summarizes the effects of visual priors, starting from a baseline (a) without the other components in this figure. These priors jointly have a gain of \textbf{27.7} points, where the canvas-based designs (c$\to$f) has a gain of \textbf{11.5} points. We discuss these components as follows.

\paragraph{2D positional embedding.} Extending from 1D positional embedding to its 2D counterpart is beneficial: see \cref{fig:visual_prior}(b)(c). 
This is observed in both (b) absolute and (c) relative positional embeddings.

To demonstrate this effect on a stronger baseline, we replace the 2D RoPE in \cref{fig:visual_prior}(f) with a 1D RoPE and observe a degradation of 3.5 points, from 54.5 to 51.0.

\paragraph{Patchification.} 
A key design principle of our method is to prepare the input as a natural image.
This enables the expansion of the token set from a very limited size (\eg, 10) to an exponentially large number. The entries \cref{fig:visual_prior}(d-f) all benefit from this design.

In \cref{fig:visual_prior}(d), we advance from 1$\times$1 patches on a 32$\times$32 canvas to 2$\times$2 patches on a 64$\times$64 canvas. Doing so does not increase the computational cost of the Transformer. In this ablation (d), the scaling ratio is fixed as 2$\times$.
As such, if we constrain each 2${\times}$2 patch to cover only one raw pixel, it becomes equivalent to the 1${\times}$1 patch counterpart on the 32${\times}$32 canvas.
Therefore, to ensure a meaningful comparison, we do not impose this constraint, allowing each $2{\times}2$ patch to cover \textit{multiple} colors. This can be interpreted as one-pixel translation augmentation on the canvas.

Even so, the 2${\times}$2 patchification leads to a noticeable gain of 2.4 points, improving from 43.0 to 45.4; see \cref{fig:visual_prior}(c,d). In spite of the small one-pixel augmentation, each patch can cover multiple colors (as in natural images), which substantially enriches the data space for learning.

% ################################################################################################################################
\begin{table}[t]
\centering
\tablestyle{6pt}{1.0}
\begin{tabular}{c|cc|rrr}
\toprule
model  & width & depth & \#params & Gflops    & acc.    \\
\midrule
\multirow{3}{*}{\textbf{ViT}}    &384 &5 & 6M   & 10 & 44.4 \\
&512 &10 & 18M  & 28 & \textbf{54.5}   \\
&768 &20 & 66M  & 99 & 53.0     \\ \midrule
\multirow{3}{*}{\textbf{U-Net}}  & \multicolumn{2}{c|}{setting (a)} & 7M    & 18        & 42.8      \\
& \multicolumn{2}{c|}{setting (b)} & 17M    & 33       & 47.5      \\
& \multicolumn{2}{c|}{setting (c)} & 55M  & 87  & 48.3  \\
\bottomrule
\end{tabular}
% \vspace{-0.5em}
\caption{\textbf{Vision backbones}. We compare variants of ViTs and U-Nets of similar sizes. U-Net settings are in appendix.
}
% \vspace{-.5em}
\label{tab:architecture_models}
\end{table}
% ################################################################################################################################

% ################################################################################################################################
\begin{figure}[t]
% \hspace{-1em}
\begin{minipage}{0.47\linewidth}
\centering
\includegraphics[height=1.0\linewidth]{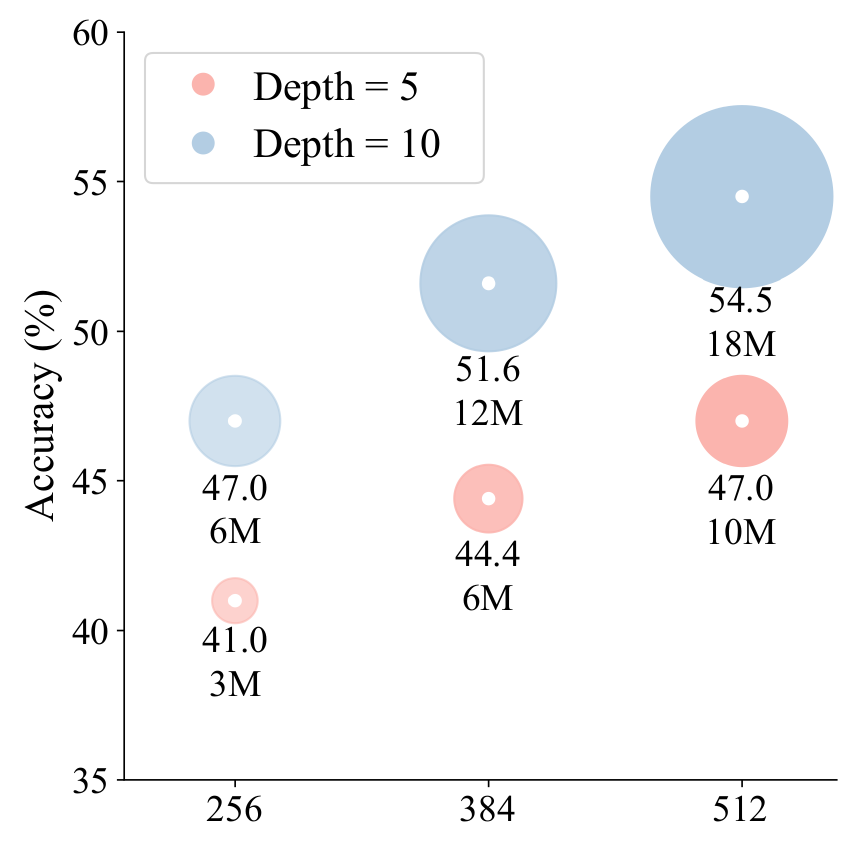}  % calibrate height
% \vspace{-2em}
\captionof{figure}{\textbf{Scalability}: ViTs with different width (x-axis) and depth. The circle areas denote model sizes.}
\label{fig:model_sizes}
\end{minipage}
\hfill
\begin{minipage}{0.47\linewidth}
\centering
\includegraphics[height=1.0\linewidth, clip, trim=0 1em 0 3em ]{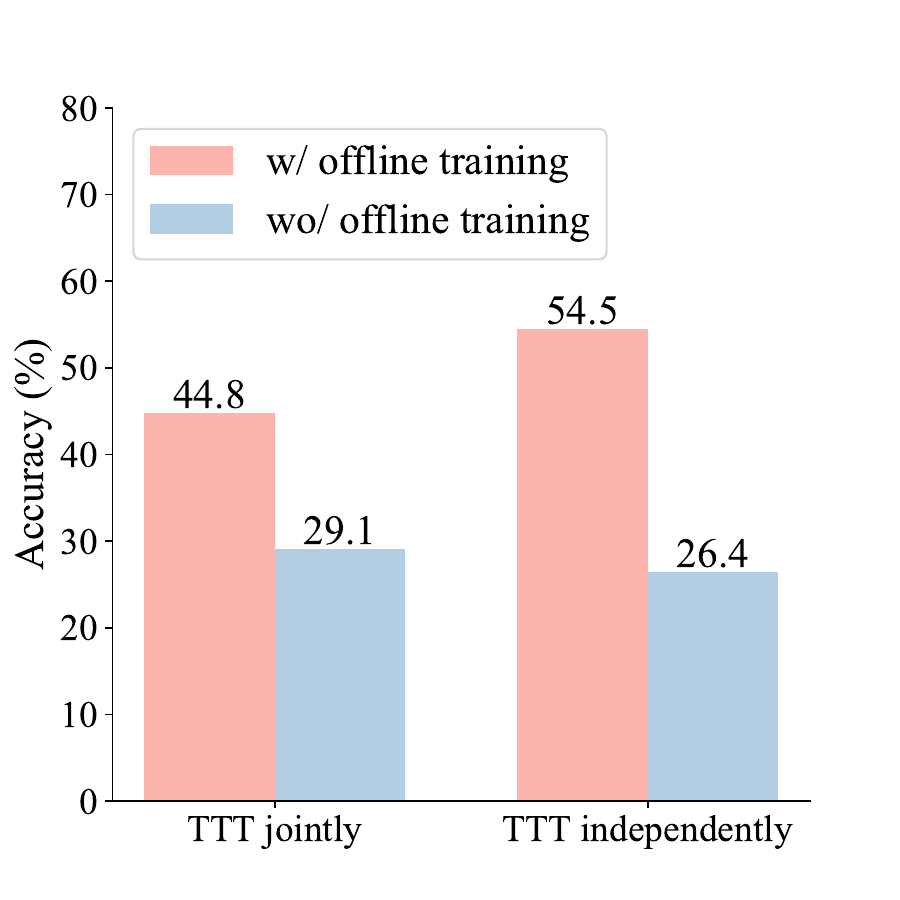}  % calibrate height
% \vspace{-2em}
\captionof{figure}{\textbf{TTT strategies}: with \vs without offline training, and joint \vs independent for each task.
}
\label{fig:ttt_ablation}
% \vspace{-2em}
\end{minipage}
\end{figure}
% ################################################################################################################################

\paragraph{Translation and scale augmentation.} In image recognition, even highly capable network architectures still benefit greatly from translation and scale augmentations. We draw similar observations in ARC. See \cref{fig:visual_prior}(e,f).

In \cref{fig:visual_prior}(e), we apply fully flexible translation augmentation on the canvas. Compared with the ``one-pixel'' augmentation in \cref{fig:visual_prior}(d), this setting yields an additional gain of 2.9 points (from 45.4 to 48.3).
In \cref{fig:visual_prior}(f), we further apply the scale augmentation enabled by the concept of canvas. Scale augmentation yields a substantial gain of \textbf{6.2} points. Unlike translation invariance, which can be partially addressed by patchification (\ie, a special form of convolution), the ViT architecture has little to no inductive bias about {scale invariance}. This can explain why scale augmentation yields a substantial gain.

% ################################################################################################################################
% ################################################################################################################################

\subsection{Other Ablation Experiments}

\paragraph{ViT \vs U-Net.} In \cref{tab:architecture_models}, we compare ViT with U-Nets, a type of convolutional network. We evaluate three model sizes for each architecture. Although ViTs consistently perform better, all U-Net variants achieve decent accuracy, suggesting that this problem can also be effectively addressed by classical vision backbones.

\paragraph{Scalability.} In \cref{fig:model_sizes}, we show ViTs with varying depths and widths. In this regime, our method demonstrates good scalability: increasing depth and/or width leads to higher accuracy as a result of better fitting. Going beyond this regime can lead to overfitting in our current setting, as shown in \cref{tab:architecture_models} for the 66M ViT model. We observe that this larger model achieves higher training accuracy, suggesting that future research should focus on generalization.

\paragraph{Test-time training (TTT) strategies.} In \cref{fig:ttt_ablation}(b), we study TTT with and without offline training, and TTT performed \textit{jointly} on all test tasks \vs \textit{independently} for each test task.

As expected, offline training greatly improves the performance of TTT, suggesting that common sense about the visual world can be learned from the training set.
We also note that even without offline training, our TTT strategy can achieve nontrivial accuracy (26.4), suggesting that some tasks in this benchmark can be solved \textit{tabula rasa}. This result outperforms that in \cite{liao2025arcagiwithoutpretraining} under a similar setting.

Surprisingly, performing TTT \textit{independently} for each test task yields substantially better performance (by $\app$10 points) than doing so \textit{jointly} across all test tasks, even though the latter relies on a stronger assumption about the availability of multiple test tasks at once.\footnotemark{} We hypothesize that overtraining on the test tasks may cause the model to forget the knowledge acquired during offline training.

\footnotetext{
In general, it cannot be assumed that multiple unseen tasks will be presented all at once.
}

\paragraph{Single-view \vs multi-view inference.} As discussed in \cref{subsec:inference}, we adopt multi-view inference by default. For completeness, we also examine the single-view inference accuracy.
Since single-view inference cannot produce multiple predictions, we compare pass@1 accuracy. See \cref{tab:single_view}.

% ################################################################################################################################
\begin{table}[t]
\centering
\tablestyle{4pt}{1.0}
\begin{tabular}{ccc}
\toprule
single-view, pass@1 & multi-view, pass@1 &  \deemph{multi-view, pass@2} \\
\midrule
35.9 & 49.8 & \deemph{54.5} \\
\bottomrule
\end{tabular}
% \vspace{-0.5em}
\caption{\textbf{Single-view \vs multi-view inference}.}
% \vspace{-1em}
\label{tab:single_view}
\end{table}
% ################################################################################################################################

Single-view inference has a decent pass@1 accuracy of 35.9; multi-view inference further boosts to 49.8, thanks to majority voting. Unlike typical computer vision applications such as semantic segmentation, in ARC, a mistake on even a single pixel renders the entire prediction incorrect. This may explain the large gain seen here.

\begin{table}[t]
\centering
\tablestyle{6pt}{1.05}
\begin{tabular}{lcrr}
\toprule
system  &  \#params  &  \textbf{ARC-1}  &  \textbf{ARC-2}  \\[0.3em]  
\multicolumn{4}{l}{\cellcolor[HTML]{EFEFEF}\textit{large language models (LLMs)}}  \\
{\deemph  Deepseek  R1}~\cite{guo2025deepseek}  &  {\deemph  671B}  &  {\deemph  15.8}  &  {\deemph  1.3}  \\
{\deemph  Claude  3.7  8k}~\cite{arcprize2025arcagi_benchmarking}  &  {\deemph  N/A}  &  {\deemph  21.2}  &  {\deemph  0.9}  \\
{\deemph  o3-mini-high}~\cite{arcprize2025arcagi_benchmarking}  &  {\deemph  N/A}  &  {\deemph  34.5}  &  {\deemph  3.0}  \\
{\deemph  GPT-5}~\cite{arcprize2025arcagi_benchmarking}  &  {\deemph  N/A}  &  {\deemph  44.0}  &  {\deemph  1.9}  \\
{\deemph  Grok-4-thinking}~\cite{arcprize2025arcagi_benchmarking}  &  {\deemph  1.7T}  &  {\deemph  66.7}  &  {\deemph  16.0}  \\
{\deemph  Bespoke  (Grok-4)}~\cite{berman2025highestscore}  &  {\deemph  1.7T}  &  {\deemph  \textbf{79.6}}  &  {\deemph  \textbf{29.4}}  \\[0.3em]  
\multicolumn{4}{l}{\cellcolor[HTML]{F3F3F3}\textit{recurrent  models}}  \\
HRM~\cite{wang2025hierarchicalreasoningmodel}  &  27M  &  40.3  &  5.0  \\
TRM~\cite{jolicoeurmartineau2025morerecursivereasoningtiny}  &  7M  &  {44.6}  &  {7.8}  \\[0.3em]  
\multicolumn{4}{l}{\cellcolor[HTML]{F3F3F3}\textit{vision  models}}  \\
\textbf{VARC}  &  18M  &  \cellcolor[HTML]{FFFFFF}\underline{54.5}  &  \cellcolor[HTML]{FFFFFF}\underline{8.3}  \\
\textbf{VARC} (ensemble)  & 73M  &\textbf{60.4}  &  \textbf{11.1}\\
[0.3em]  
\multicolumn{4}{l}{\cellcolor[HTML]{EFEFEF}\textit{human  results}}  \\
avg. human~\cite{legris2024harcrobustestimatehuman}  &  -  &  60.2  &  -  \\
best human~\cite{arcprize2025arcagi_benchmarking}  &  -  &  98.0  &  100.0  \\  \bottomrule
\end{tabular}
% \vspace{-.5em}
\caption{\textbf{System-level comparisons} on the ARC-1 and ARC-2 benchmarks. LLM-based results are from the ARC-AGI leaderboard  \cite{arcprize2025arcagi_benchmarking}. HRM, TRM, and our VARC are trained from scratch only on ARC data. 
Our single-model result is based on ViT, with \mbox{mean$\pm$std} of 54.5$\pm$0.7 (ARC-1) and 8.3$\pm$0.4 (ARC-2) over four runs.
Our ensemble result aggregates an 18M ViT and a 55M \mbox{U-Net}, each with test-time training performed four times.}
\label{tab:arc-final}
\vspace{-0.8em}
\end{table}

\subsection{System-level Comparisons}

In \cref{tab:arc-final} we compare with leading results using LLMs or recurrent models, on ARC-1 and ARC-2.\footnotemark 

Our model compares favorably with some of the most powerful LLMs at the time their results were reported: including Deepseek, Claude, o3, and GPT-5 (we note that given the rapid progress of LLMs, these models may have stronger results by the time our paper is public). LLMs are pre-trained on internet-scale data, and some may also incorporate multimodal data that include images. Our method does not rely on such data and uses a model that is several orders of magnitude smaller.

In the \textit{controlled} setting of training from scratch on ARC data, our method substantially outperforms the recurrent models: HRM \cite{wang2025hierarchicalreasoningmodel} and TRM \cite{jolicoeurmartineau2025morerecursivereasoningtiny}. Our VARC with 18M parameters is $\app$10 points better than TRM on ARC-1, a $>$20\% relative improvement. Note that, once test-time training is completed, our model performs fully \textit{feedforward} inference, with \textit{no recurrence} involved in reasoning.

Following the classical ensembling practice in vision (\eg, AlexNet \cite{krizhevsky2012imagenet}), we ensemble one ViT and one U-Net, each with test-time training run four times.
Doing so boosts our result to \textbf{60.4}. This result closes the gap with the reported average human performance (60.2 \cite{legris2024harcrobustestimatehuman}).

\footnotetext{Our ARC-2 models are trained only on the ARC-1 dataset, with test-time training and inference on the ARC-2 set.}

\section{Visualization and Analysis}

% ################################################################################################################################
\begin{figure}[t]
\centering
\includegraphics[width=1.0\linewidth,clip,trim={1.4cm 5cm 10.4cm 1cm}]{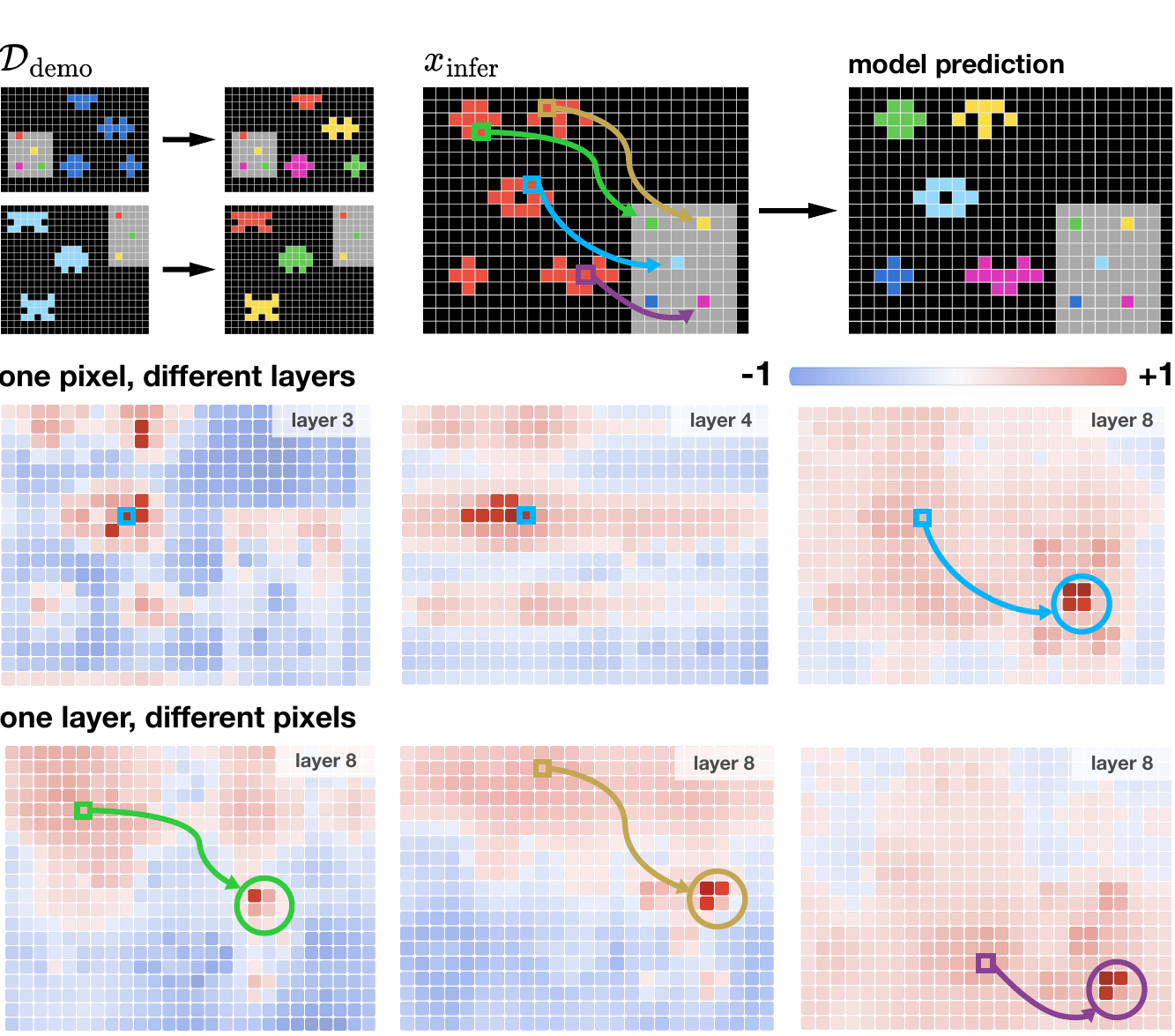}
\caption{
\textbf{Visualization of pixel-to-pixel attention.}
\textbf{(Top)}: a test task from ARC-1 eval: showing demo pairs, inference input, and model prediction.
\textbf{(Middle)}: attention maps for a single pixel across different layers.
With the highlighted pixel as query, we show pre-softmax logits.
Different layers exhibit different behavior.
\textbf{(Bottom)}:
attention maps in layer 8 with other query pixels.
All of them correctly attend to their corresponding palette pixel. 
}
\vspace{-0.5em}
\label{fig:attenion_pixel_lavel}
\end{figure}
% ################################################################################################################################

Beyond numerical metrics, we provide additional qualitative results that help reveal the model’s behavior. We refer readers to the appendix for more visualizations.

\paragraph{Attention patterns.} \cref{fig:attenion_pixel_lavel} shows the attention patterns of our ViT model in a test task. These attention maps show that our model can correctly reason about the relationship between a source pixel and its target pixel to copy from. 

\begin{figure}[t]
\centering
\includegraphics[width=0.95\linewidth]{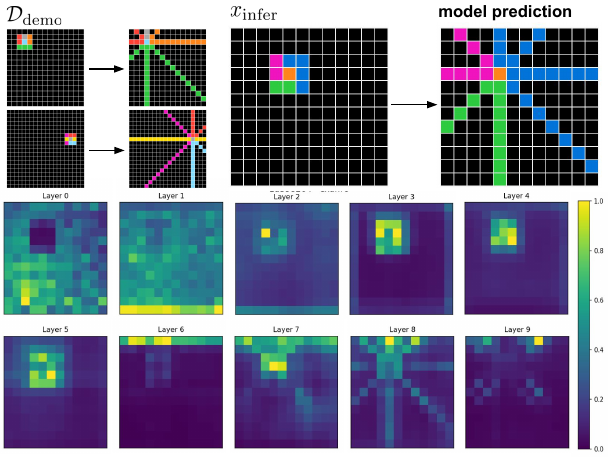}
\caption{
\textbf{Visualization of layer-wise attention maps.}
For each layer, we compute pixel-to-pixel attention and then average the softmax maps across all pixels to obtain a single map per layer. This map reveals which pixels are most attended in this layer.
We show a test task from ARC-1 eval.
In this task, some layers exhibit strong attention to the $3\times 3$ neighborhood, reflecting the influence of the pattern’s core. In comparison, some other layers (e.g., layers 7–9) focus on the outward-radiating rays, corresponding to the rule that extends colored pixels along the eight directions.
}
% \vspace{-0.3em}
\label{fig:heatmap_avg}
\end{figure}

\Cref{fig:heatmap_avg} visualizes the layer-wise attention maps for another test task. A layer-wise map is the softmax attention map averaged across all pixels in the layer: it reveals which pixels receive the most attention in that layer. In this task, different layers exhibit different specialties: some layers attend to the pixels that are to be copied, and some layers attend to the target lines alone the eight directions.

\begin{figure}
\centering
\includegraphics[width=1.0\linewidth]{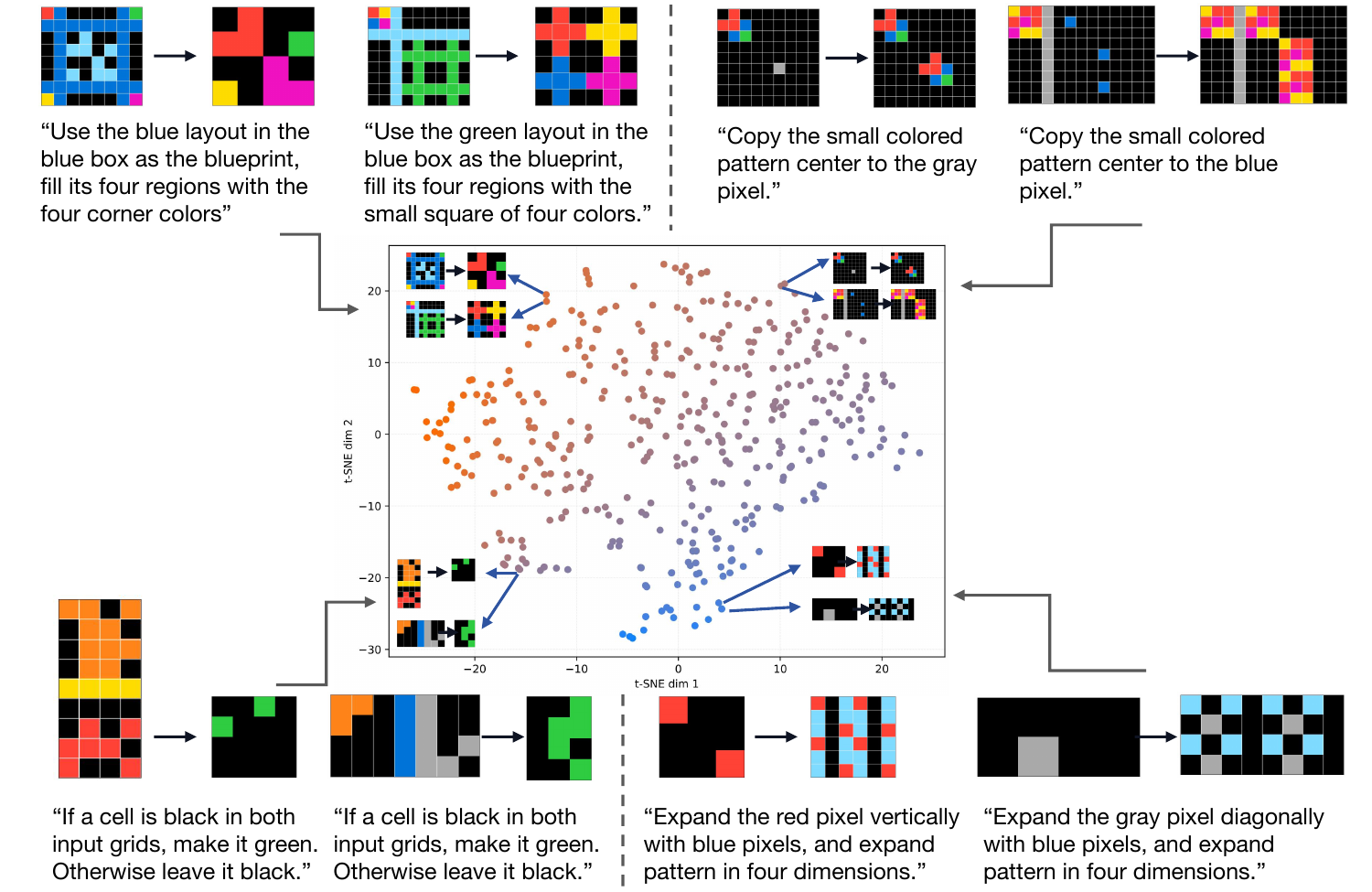}
\caption{
\textbf{t-SNE of task embeddings}, on the 400 task tokens learned from the ARC-1 training set. Each point represents a single task.
To aid the reader, we provide human-written descriptions for the tasks (which are not used in any form by our method).
}
\label{fig:task_token_tsne}
\vspace{-0.8em}
\end{figure}

\paragraph{t-SNE of task embeddings.} Our model is conditioned on a task token, with an embedding learned to represent each task. With 400 training tasks in ARC-1, our model learns 400 distinct task embeddings in offline training. We visualize these 400 embeddings in the 2D space by t-SNE \cite{maaten2008visualizing} (see \cref{fig:task_token_tsne}). \textit{Each point corresponds to a task.}.

Interestingly, we observe that nearby points in the task embedding space exhibit similar semantics. For example, the top-left corner in \cref{fig:task_token_tsne} shows two  tasks related to coloring; the bottom-left corner shows two tasks related to generalized logic operations (\ie, AND/OR/XOR).
This visualization suggests that our method attempts to learn the \textit{relations} between different tasks, which is an essential ability for abstraction and reasoning.

% ################################################################################################################################
% ################################################################################################################################

% ################################################################################################################################
% ################################################################################################################################

\section{Conclusion}

Our work explores a previously overlooked perspective in the ARC task by framing it as an image-to-image translation problem. It naturally enables the adaptation of visual frameworks and yields strong few-shot generalization competitive with recent approaches, while remaining orders of magnitude smaller than most LLM-based models. This opens up a new possibility of treating ARC as a vision-centric problem, emphasizing abstraction and reasoning emerging directly from image pixels.

We hope this work will encourage the community to leverage ARC not only as a symbolic reasoning problem, but also as a testbed for promoting the generalization capacity of visual methods. 
Future research may extend this direction through more expressive architectures, richer visual priors, or larger-scale image pre-training. We envision that  vision-centric reasoning will play a key role in building AI systems capable of learning and applying abstract concepts in a human-like manner.

\small
\bibliographystyle{ieeenat_fullname}
\bibliography{main}
\appendix
\clearpage
% \setcounter{page}{1}
% \maketitlesupplementary

\section{Additional Implementation Details}

\subsection{Configurations}

We report the training configurations in~\cref{tab:hyper_pipeline}. The running time under this configuration is profiled in~\cref{tab:train_time_estimate}. 

The hyperparameters for our ViT models are listed in~\cref{tab:hyper_arc_vit_architecture}, and those for our U-net models are shown in~\cref{tab:unet_architecture_details}. 

\begin{table}[t]
\centering
\small
\tablestyle{6pt}{1.05}
\begin{tabular}{lr}
\toprule
\multicolumn{2}{l}{\cellcolor[HTML]{EFEFEF}\textit{\textbf{offline training}}}                            \\
\multicolumn{1}{l|}{epochs}                                                    & 100                      \\
\multicolumn{1}{l|}{warmup epochs}                                             & 10                       \\
\multicolumn{1}{l|}{optimizer}                                                 & Adam~\cite{adam2014method}, betas=(0.9, 0.999)                    \\
\multicolumn{1}{l|}{batch size}                                                & 32                       \\
\multicolumn{1}{l|}{learning rate}                                             & 3e-4                        \\
\multicolumn{1}{l|}{learning rate scheduler}                                   & cosine                   \\
\multicolumn{1}{l|}{weight decay}                                              & 0                        \\
\multicolumn{1}{l|}{dropout}                                                   & 0.1                      \\ 
% \multicolumn{1}{l|}{data points num}                                                   & $\sim$400$\times$1k                      \\ 
\midrule
\multicolumn{2}{l}{\cellcolor[HTML]{EFEFEF}\textit{\textbf{test-time training}}}                  \\
\multicolumn{1}{l|}{epochs}                                                    & 100                      \\
\multicolumn{1}{l|}{warmup epochs}                                             & 10                       \\
\multicolumn{1}{l|}{optimizer}                                                 & Adam~\cite{adam2014method}, betas=(0.9, 0.999)                    \\
\multicolumn{1}{l|}{batch size}                                                & 8                        \\
\multicolumn{1}{l|}{learning rate}                                             & 3e-4                        \\
\multicolumn{1}{l|}{learning rate scheduler}                                   & cosine                   \\
\multicolumn{1}{l|}{weight decay}                                              & 0                        \\
\multicolumn{1}{l|}{dropout}                                                   & 0.1                      \\                
% \multicolumn{1}{l|}{data points num}                                                   & $\sim$3$\times$60                      \\ 
% \midrule
% \multicolumn{2}{l}{\cellcolor[HTML]{EFEFEF}\textit{\textbf{test-time training and inference aug.}}}                                \\
% \multicolumn{1}{l|}{flip}                                                      & horizontal, vertical \\
% \multicolumn{1}{l|}{rotation}                                                  & 0$^\circ$, 90$^\circ$, 180$^\circ$, 270$^\circ$          \\
% \multicolumn{1}{l|}{color permutations}                                         & 10 (including original)                        \\
% \multicolumn{1}{l|}{num attempt*} & 10    \\ 
\bottomrule     
\end{tabular}
\vspace{-.5em}
\caption{\textbf{Configurations}.}
\label{tab:hyper_pipeline}
\end{table}

\begin{table}[t]
\centering
\tablestyle{6pt}{1.05}
\begin{tabular}{lr}
\toprule
\multicolumn{2}{l}{\textit{\textbf{offline training}}}                            \\
\midrule
\multicolumn{1}{l|}{GPU type}                                                   & H100                      \\ 
\multicolumn{1}{l|}{GPU number}                                                   & 8                      \\ 
\multicolumn{1}{l|}{GPU time}                                                   & 4.8 hours                      \\ 
\bottomrule
\end{tabular}
\hspace{2em}
\begin{tabular}{lr}
\toprule
\multicolumn{2}{l}{\textit{\textbf{test-time training}}}                  \\
\midrule
\multicolumn{1}{l|}{GPU type}                                                   & H100                      \\ 
\multicolumn{1}{l|}{GPU number}                                                   & 1                      \\ 
\multicolumn{1}{l|}{GPU time}                                                   & 0.7s per epoch                    \\ 
\bottomrule     
\end{tabular}
\caption{\textbf{Running time} of the ViT-18M model. The reported time is obtained with \texttt{torch.compile} optimization.}
\label{tab:train_time_estimate}
\end{table}

\begin{table}[t]
\centering
\tablestyle{6pt}{1.05}
\begin{tabular}{l|ccc}
\toprule
{\textbf{ViT}} & {\textbf{6M}} & {\textbf{18M}} & {\textbf{66M}} \\
\midrule
hidden dim                  & \multicolumn{1}{c}{384}                                           & 512                   & 768                   \\
Transformer blocks          & \multicolumn{1}{c}{5}                                             & 10                    & 20                    \\
\# heads             & \multicolumn{1}{c}{8}                                             & 8                     & 12                    \\
MLP block hidden dim       & \multicolumn{3}{c}{512}                                                                                           \\
dropout                    & \multicolumn{3}{c}{0.1}                                                                                           \\
% positional embed           & \multicolumn{3}{c}{2D RoPE (to query/key)}                                                                      \\
patch size                 & \multicolumn{3}{c}{2$\times$2}                                                                                           \\
canvas size          & \multicolumn{3}{c}{64$\times$64}                                                                                         \\ \bottomrule
\end{tabular}
\caption{\textbf{Configuration of the ViT architecture}. The 18M model is our default setting.}
\label{tab:hyper_arc_vit_architecture}
\end{table}

\begin{table}[t]
\centering
\tablestyle{6pt}{1.05}
\begin{tabular}{l|rrr}
\toprule
\textbf{U-net} & {\textbf{7M}} & {\textbf{17M}} & {\textbf{55M}} \\ 
\midrule
\# stages      & 3                     & 3                     & 3                     \\
layers per stage              & 1                     & 1                     & 2                     \\
\# channels at resolution 1         & 80                    & 120                   & 160                   \\
attention at resolution 1       & No                    & No                    & No                    \\
\# channels at resolution 2         & 160                   & 240                   & 320                   \\
attention at resolution 2       & Yes                   & Yes                   & Yes                   \\
\# channels at resolution 3         & 160                   & 240                   & 320                   \\
attention at resolution 3       & Yes                   & Yes                   & Yes                   \\
mid block               & No                    & No                    & Yes                   \\ \bottomrule
\end{tabular}
\caption{\textbf{Configuration of the U-Net architecture}. The definition follows standard U-Nets used in generative models \cite{Song2019,dhariwal2021diffusion}.}
\label{tab:unet_architecture_details}
\end{table}

\subsection{Test-time Training Augmentation}

During test-time training, we augment the single test task $T$ into multiple \textit{auxiliary} tasks. We use a distinct task embedding for each auxiliary task, as not all of these augmentations correspond to the same underlying rule (\eg, consider ``gravity'' under a 90$^\circ$ rotation). 
We apply 2 flippings (horizontal and vertical) or 3 rotations (in multiples of 90$^\circ$), and 10 predefined color index permutations, resulting in $(2{+}3){\times}10{=}50$ auxiliary tasks with the original task.
We train for 100 epochs on these 51 tasks, covering $100 \times 51 \times 3 = 15.3$k samples in total for test-time training for one test task $T$ (assuming 3 raw samples in this task).

\subsection{Shape Handling}

\begin{figure}
    \centering
    \includegraphics[width=1.0\linewidth]{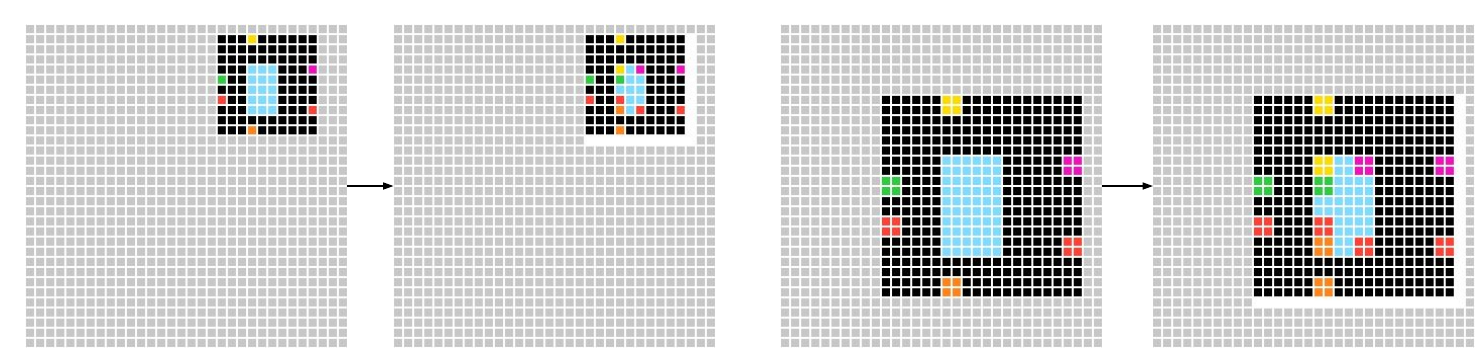}
    \caption{\textbf{Shape Handling.} The gray pixels denote the background tokens \texttt{[BG]}, which keep the canvas size fixed (64$\times$64 by default). The white pixels denote the border tokens \texttt{[BD]}, which indicate the output shape. (Left): a pair $(x, y)$ with a scaling ratio of $1{\times}$. (Right): a pair $(x, y)$ with a scaling ratio of $2{\times}$.}
    \label{fig:predict_shape_illustration}
\end{figure}
\label{sec:shape_pred}

Unlike standard semantic segmentation, in ARC, the \textit{raw} input and output sizes are not always identical (\eg, see \cref{fig:train_protocal}, Test Set, Task 1).
This issue can be addressed on the canvas in a unified framework. In our method, the input/output canvas \textit{always} has a fixed size and is filled with a background token \texttt{[BG]}. In addition, when the raw output is placed on the canvas (serving as the ground truth during training), we always use an extra border token, \texttt{[BD]}, to indicate the right and bottom edges. Specifically, the token \texttt{[BD]} is filled along the one-pixel-wide edge on the right and bottom sides. During inference, we locate the rightmost and bottommost \texttt{[BD]} tokens and crop the output accordingly to recover the final predicted shape. This is illustrated in~\cref{fig:predict_shape_illustration}.

Since the number of background pixels \texttt{[BG]} can dominate in some examples, we apply \textit{attention masks} in the self-attention blocks to encourage the model to focus on the foreground pixels.
The attention masks are applied after the query-key dot-product computation, adding a large negative value to the keys corresponding to background inputs. The resulting softmax attention scores are therefore zero at those key positions. Moreover, during training, the loss is computed only on locations where the inputs are not background pixels \texttt{[BG]}. These designs encourage the model to pay more attention to foregrounds and therefore improve accuracy, although we note that even without them, our method still performs competitively, as observed in our preliminary experiments.

\section{Additional Experiments}

\subsection{Offline Training Data Scaling}
\begin{figure}
    \centering
    \includegraphics[width=0.6\linewidth]{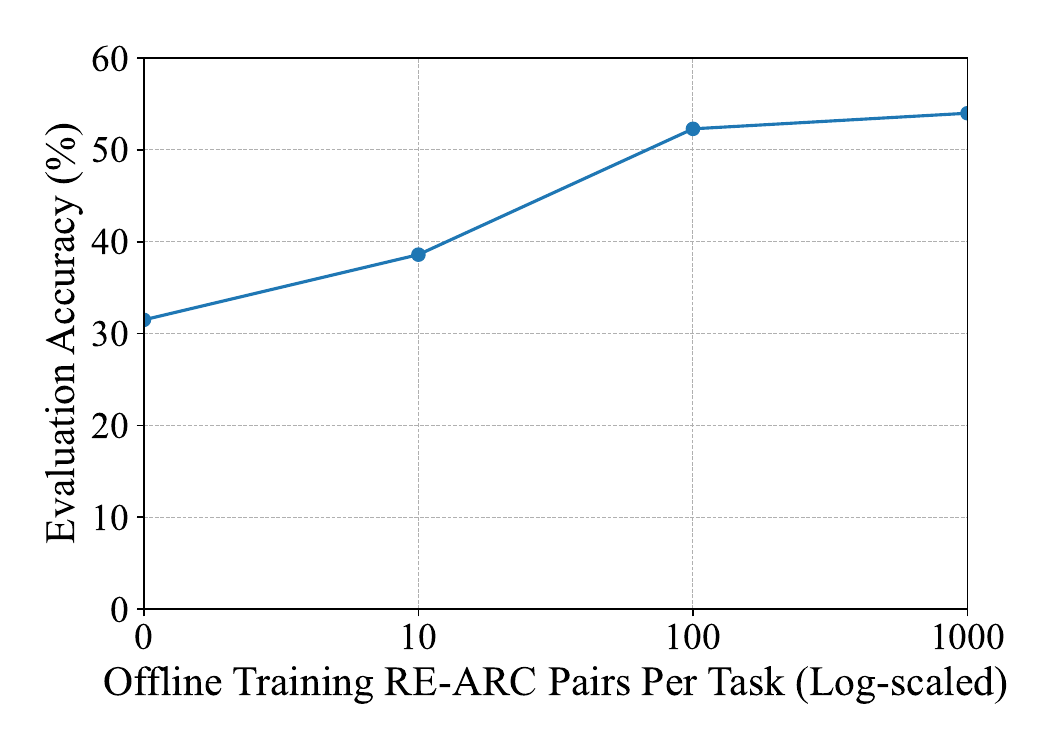}
    \vspace{-1em}
    \caption{\textbf{Offline training data scaling:} effect of varying the number of RE-ARC samples per task, evaluated on the ARC-1 eval set. Increasing the amount of offline training data is beneficial, although even without it, our model can achieve decent accuracy. 
    }
    \vspace{-1em}
    \label{fig:pretrain_scale}
\end{figure}

\begin{figure}
    \centering
    \includegraphics[width=0.6\linewidth]{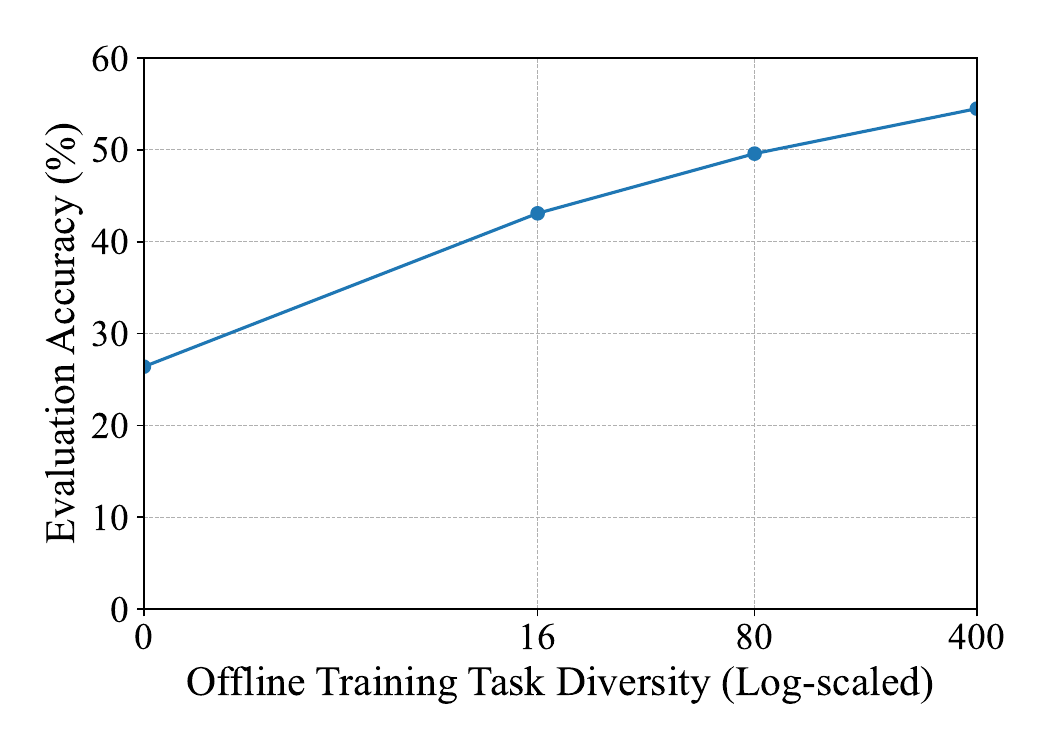}
    \vspace{-1em}
    \caption{\textbf{Offline training task diversity scaling:} effect of varying the number of training tasks, evaluated on the ARC-1 eval set. Increasing task diversity is beneficial. }
    \vspace{-1em}
    \label{fig:task_diversity}
\end{figure}
Since we use the RE-ARC dataset~\cite{hodel2024addressingabstractionreasoningcorpus} in our offline training, we can examine the effect of \textit{data scale provided by RE-ARC}. See \cref{fig:pretrain_scale}. Using only the original ARC training data, \textit{without} any RE-ARC data, our method achieves a decent accuracy of 31.5.
By adding 10, 100, and 1,000 pairs per task from RE-ARC, the accuracy increases to 38.6, 52.3, and 54.0, respectively. This comparison suggests that increasing the amount of offline training data is beneficial, although the returns diminish beyond a certain point. 

Beyond scaling the data per task using RE-ARC, we also examine the scalability of the \textit{offline training task diversity}. See \cref{fig:task_diversity}.
When trained on 0, 16, 80, and 400 tasks, the accuracy increases from 26.4 to 43.1, 49.6, and 54.5, respectively, suggesting that the diversity of training tasks is helpful for generalization.

\begin{figure}[t]
  \centering
  \includegraphics[width=0.49\linewidth]{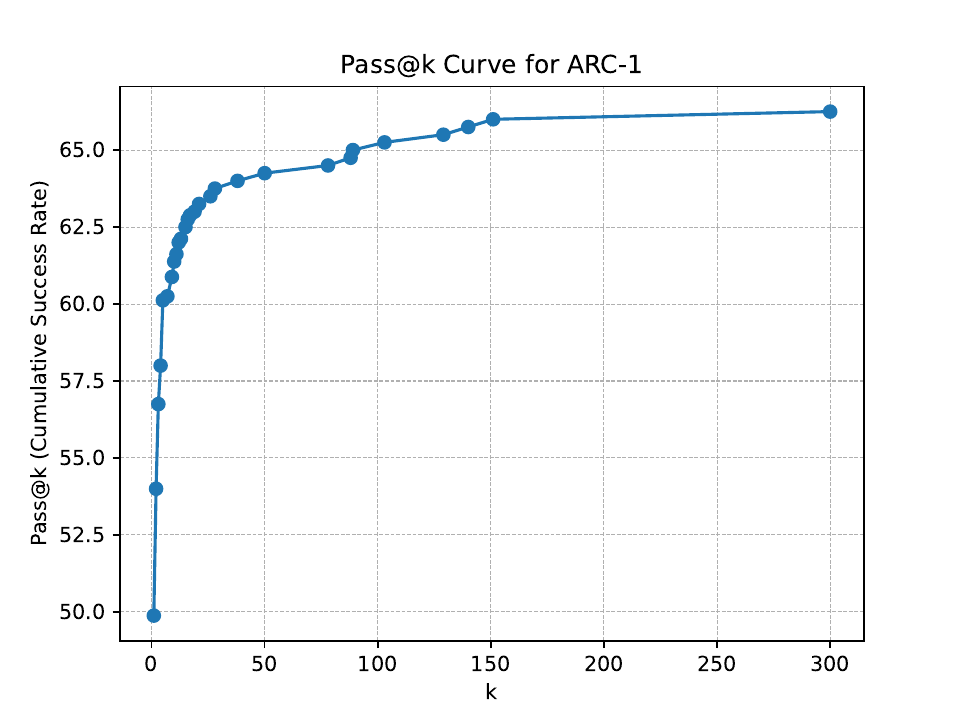}
  \includegraphics[width=0.49\linewidth]{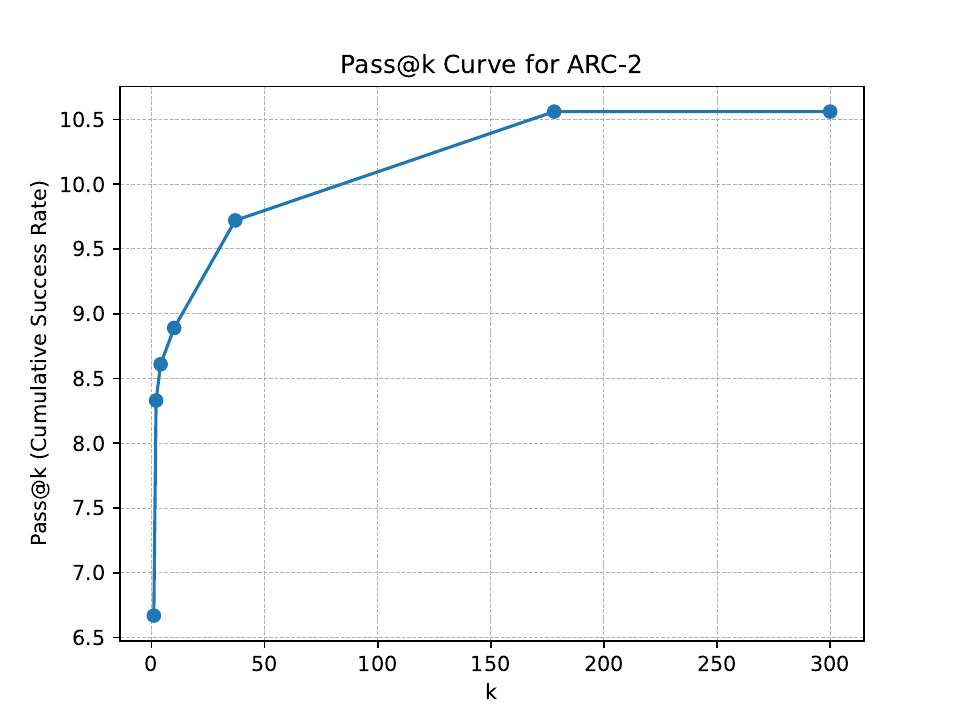}
  \\
  \includegraphics[width=0.49\linewidth]{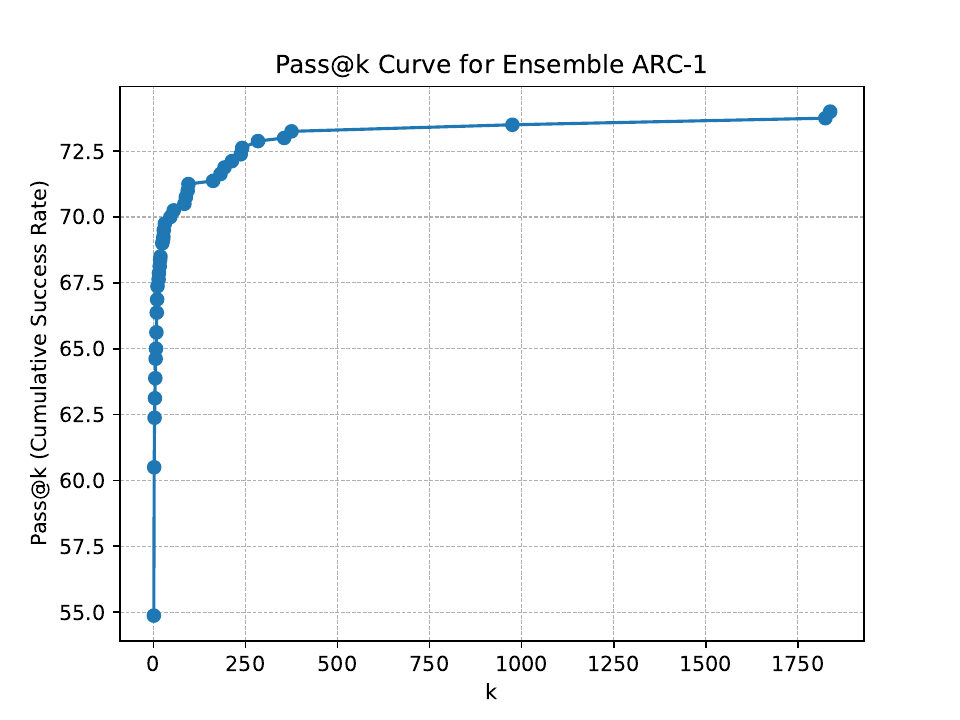}
  \includegraphics[width=0.49\linewidth]{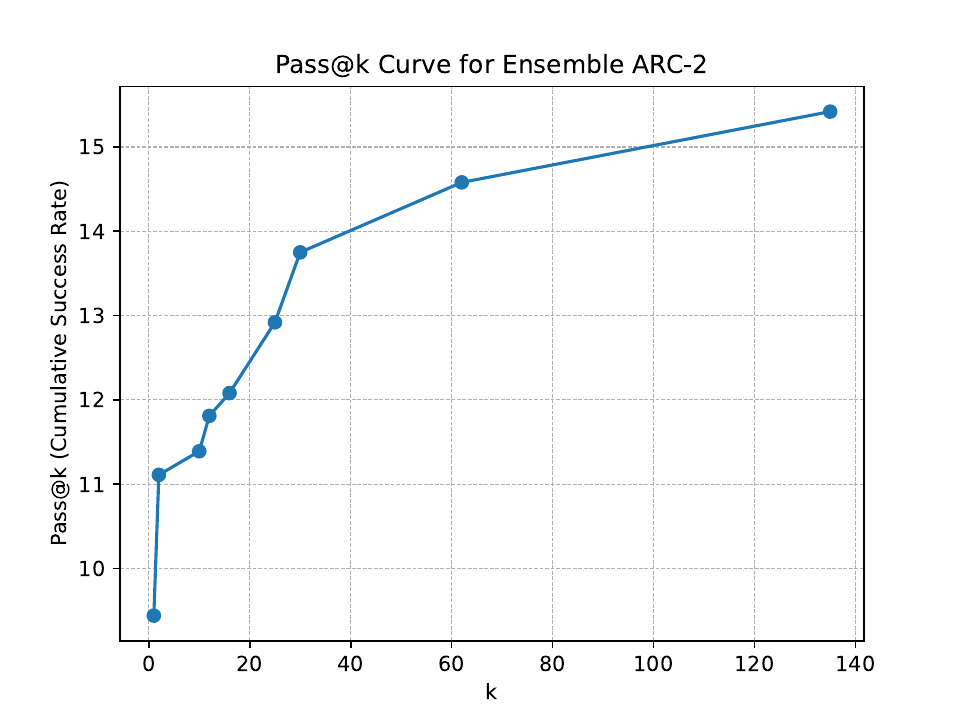}
  \\
  \vspace{-1em}
  \caption{\textbf{Pass@k results} in the ARC-1 (left) and ARC-2 (right) evaluation sets. Results are obtained with majority voting from multi-view inference, using 510 views. 
  (Top): using a single model of ViT-18M. (Bottom): using an ensemble of one ViT-18M and one U-Net-55M, each with test-time training run four times.
  }
  \label{fig:passk_agi1}
  \vspace{-1em}
\end{figure}

\subsection{Pass@k Results}

By default, the ARC protocol evaluates the pass@2 accuracy. We further examine the pass@$k$ accuracy, thanks to our multi-view inference with many views (510). This metric reflects whether at least one of the $k$ predicted solutions is correct. It can be viewed as a \textit{recall}-like measure.

\Cref{fig:passk_agi1} provides the pass@k results on ARC-1 and ARC-2 eval sets. As expected, as the number of proposals ($k$) increases, the pass@$k$ accuracy increases. On ARC-1, the pass@k accuracy is 49.8, 54.5, and 66.3, when $k$ is 1, 2, and 300, respectively (\cref{fig:passk_agi1}, top-left).
This result indicates that our model produces correct predictions in some of the many views, although such correct cases are not sufficiently populated to be retained after voting.
On the other hand, this result reveals the upper-bound performance (66.3) of our method, even if oracle voting were applied. Beyond voting, future efforts should focus on improving the fundamental ability of the model on each individual view.

\section{Additional Visualizations}

\subsection{Successful and Failed Examples}

\begin{figure*}
  \centering
  \includegraphics[width=0.92\linewidth]{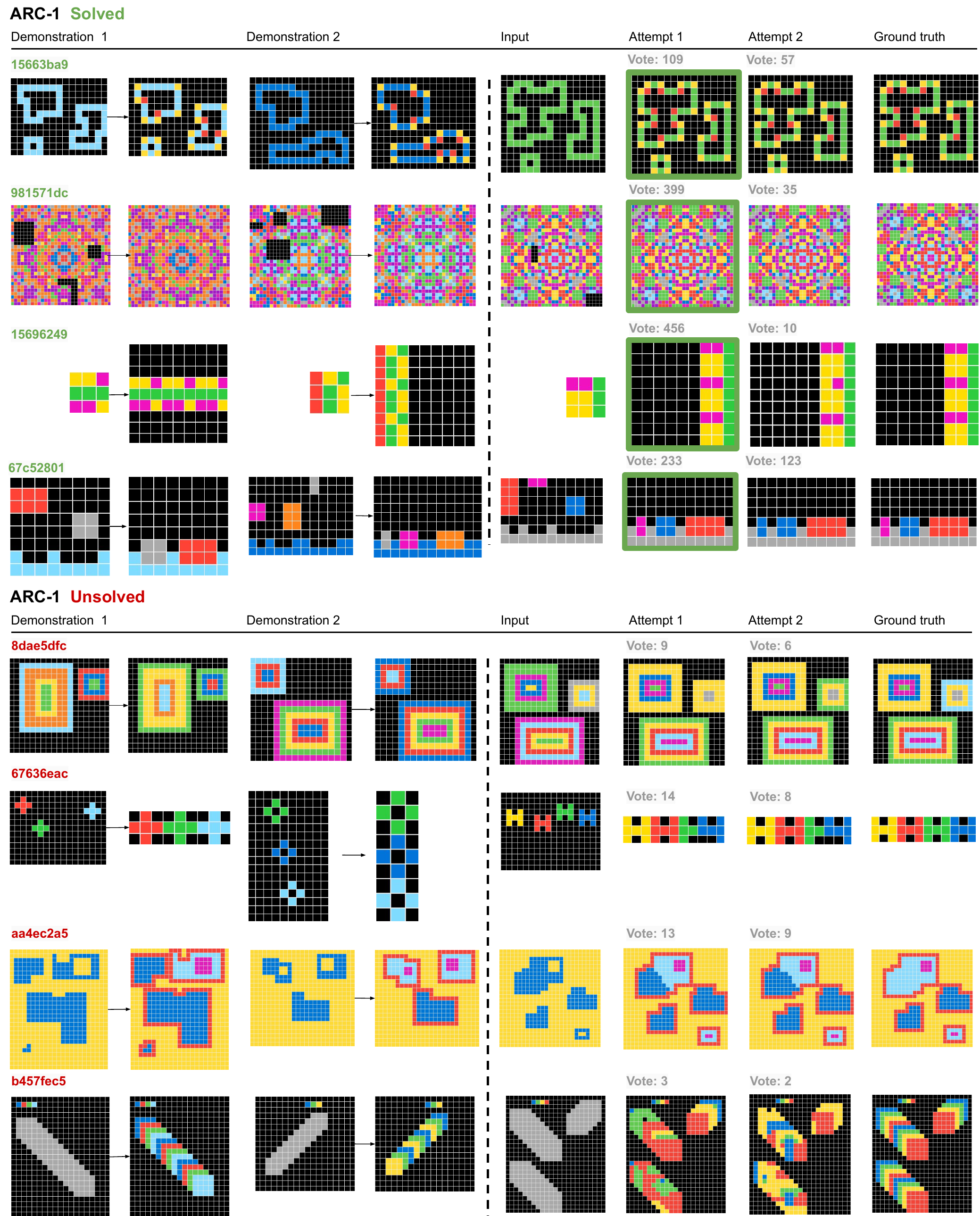}
  \caption{\textbf{Successful and failed examples on ARC-1}. \textbf{(Top)}: Examples of test tasks successfully solved by VARC. \textbf{(Bottom)}: Examples of test tasks unsolved by VARC. \textbf{(Left)}: Two demonstration example pairs shown for each task (some have more demonstrations not shown here). \textbf{(Right)}: Inference input and the first and second solutions proposed by VARC. The green box indicates the correct output.
  }
  \label{fig:solved_arc1}
\end{figure*}

\begin{figure*}
  \centering
  \includegraphics[width=0.95\linewidth]{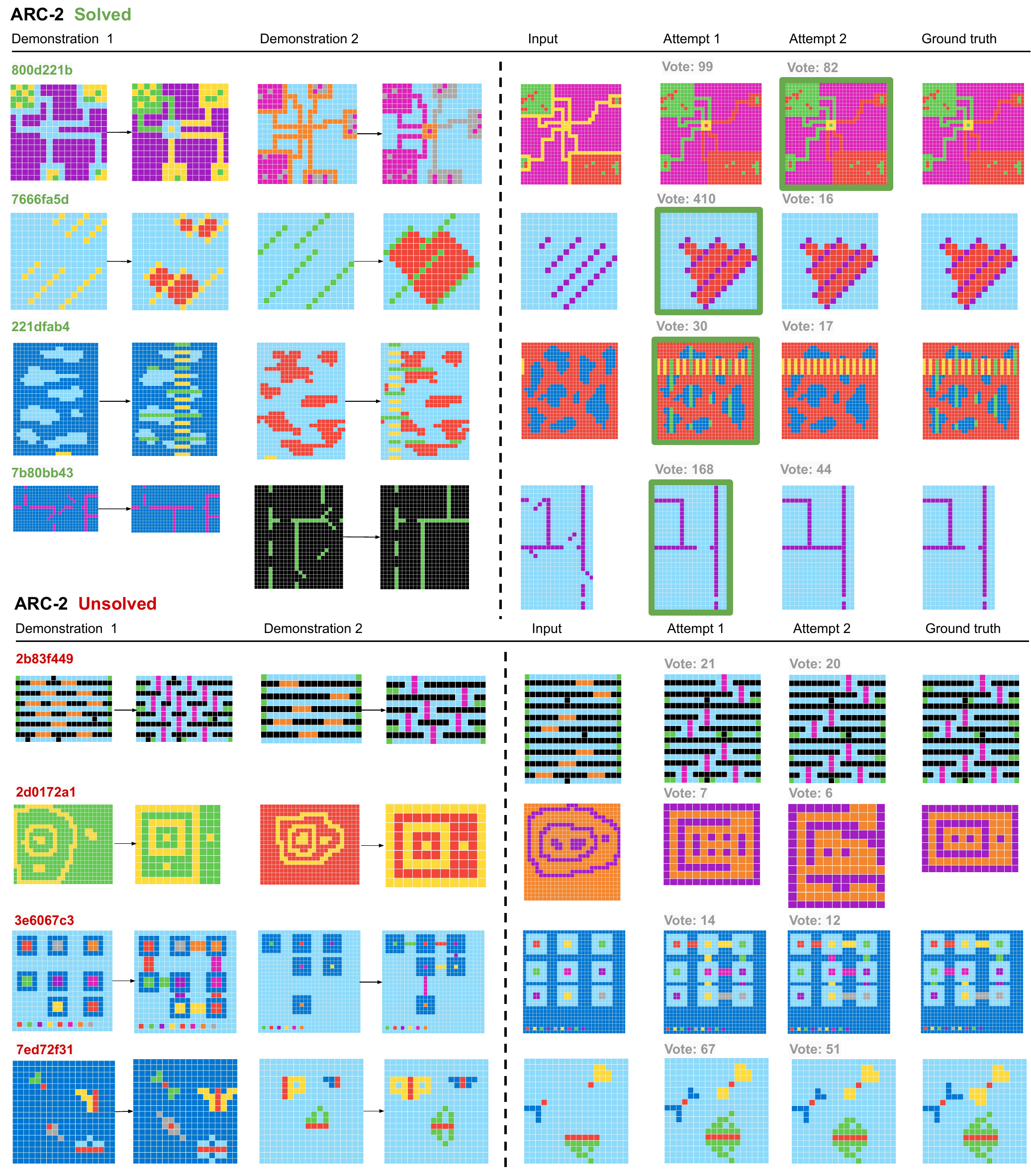}
  \caption{\textbf{Successful and failed examples on ARC-2}. \textbf{(Top)}: Examples of test tasks successfully solved by VARC. \textbf{(Bottom)}: Examples of test tasks unsolved by VARC. \textbf{(Left)}: Two demonstration example pairs shown for each task (some have more demonstrations not shown here). \textbf{(Right)}: Inference input and the first and second solutions proposed by VARC. The green box indicates the correct output. 
  }
  \label{fig:solved_arc2}
\end{figure*}

We show successful and failed examples on ARC-1 (\cref{fig:solved_arc1}) and ARC-2 (\cref{fig:solved_arc2}).
See captions for detailed descriptions. Our method can solve some highly challenging tasks, but still makes mistakes on some tasks that are simple for humans.

\subsection{Ambiguous Examples.}

\begin{figure*}
  \centering
  \vspace{-1em}
\includegraphics[width=0.85\linewidth]{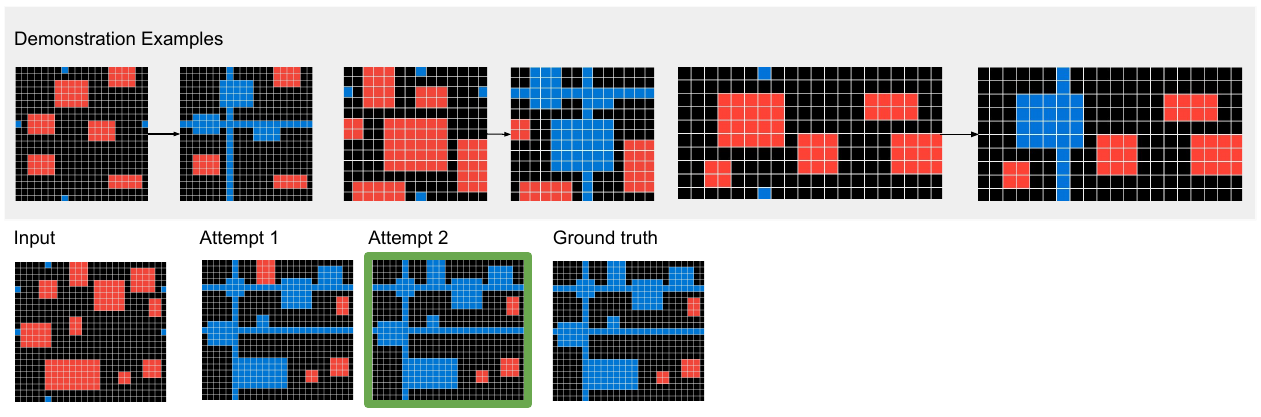}
  \caption{\textbf{Ambiguous examples}. Although most ARC tasks are unambiguous, some may admit multiple plausible explanations or rules. Here, in the given three demonstration examples of a test task (top panel), it is unclear whether a blue line ``\textit{touching}'' (but not ``\textit{going through}'') a red rectangle should render that rectangle blue. 
  The inference example (bottom panel) involves this situation (``touching''), and our model attempts to interpret the rule as either ``going-through-only'' (attempt 1) or ``touching'' (attempt 2).
  }
  \label{fig:ambinuous}
\end{figure*}

Although most ARC tasks are unambiguous, some may admit multiple plausible explanations or rules.
We show an example in \cref{fig:ambinuous}, in which our method uncovers different solutions that are plausible.
Here, the rule can be interpreted as either ``turn the red box blue only if the extended blue lines \textit{go through} the box'' (our method’s first guess) or ``turn the red box blue if the extended blue lines \textit{touch} the box in any form'' (our method’s second guess).

\subsection{Attention Maps}

\begin{figure*}
  \centering
\vspace{-1em}
\includegraphics[width=0.9\linewidth,trim={0.1cm 17.8cm 6cm 0cm},clip]{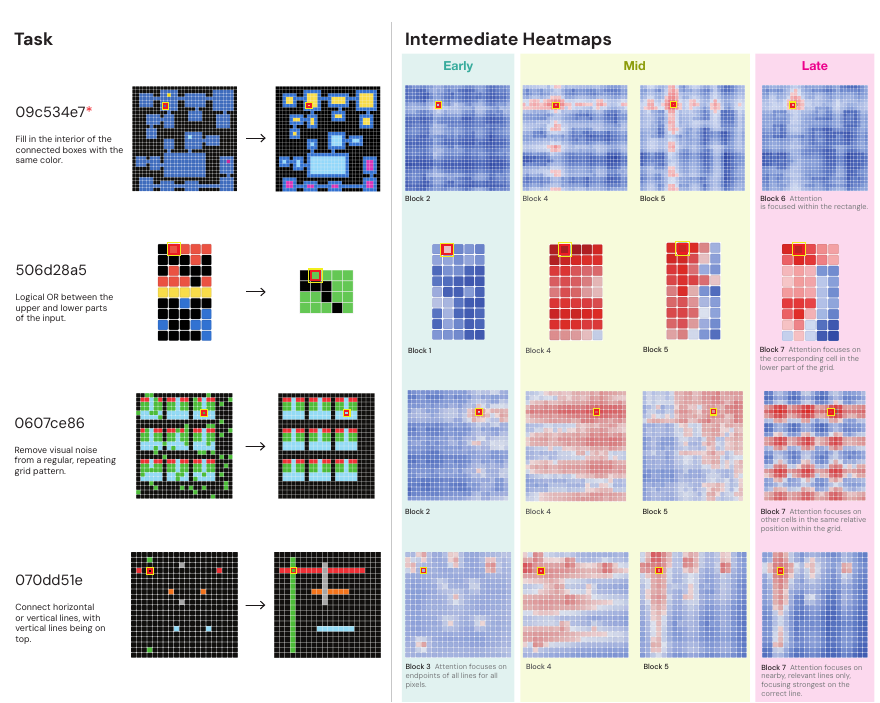}
  \caption{\textbf{Additional visualization: pixel-level attention maps}. The maps are shown for different Transformer blocks, with a query pixel highlighted by a red-yellow border. Here we show 4 test tasks in ARC eval. Layers at different depths tend to focus on different structures.
  Early layers tend to focus on local transformations and context.
  Middle layers tend to perform a more non-local connection, \eg, horizontally or vertically.
  The deep layers are more task-specialized. The red asterisk indicates the task that was not  correctly solved.
  (\textit{Here, the text descriptions are written by humans solely to help readers interpret the tasks.})
  }
  \label{fig:heatmap1}
\end{figure*}

\paragraph{Pixel-wise Attention Maps.} In~\cref{fig:heatmap1}, we visualize the attention maps of a single pixel specified as the query. See captions for detailed descriptions.

\begin{figure*}
  \centering
  \includegraphics[width=1.0\linewidth]{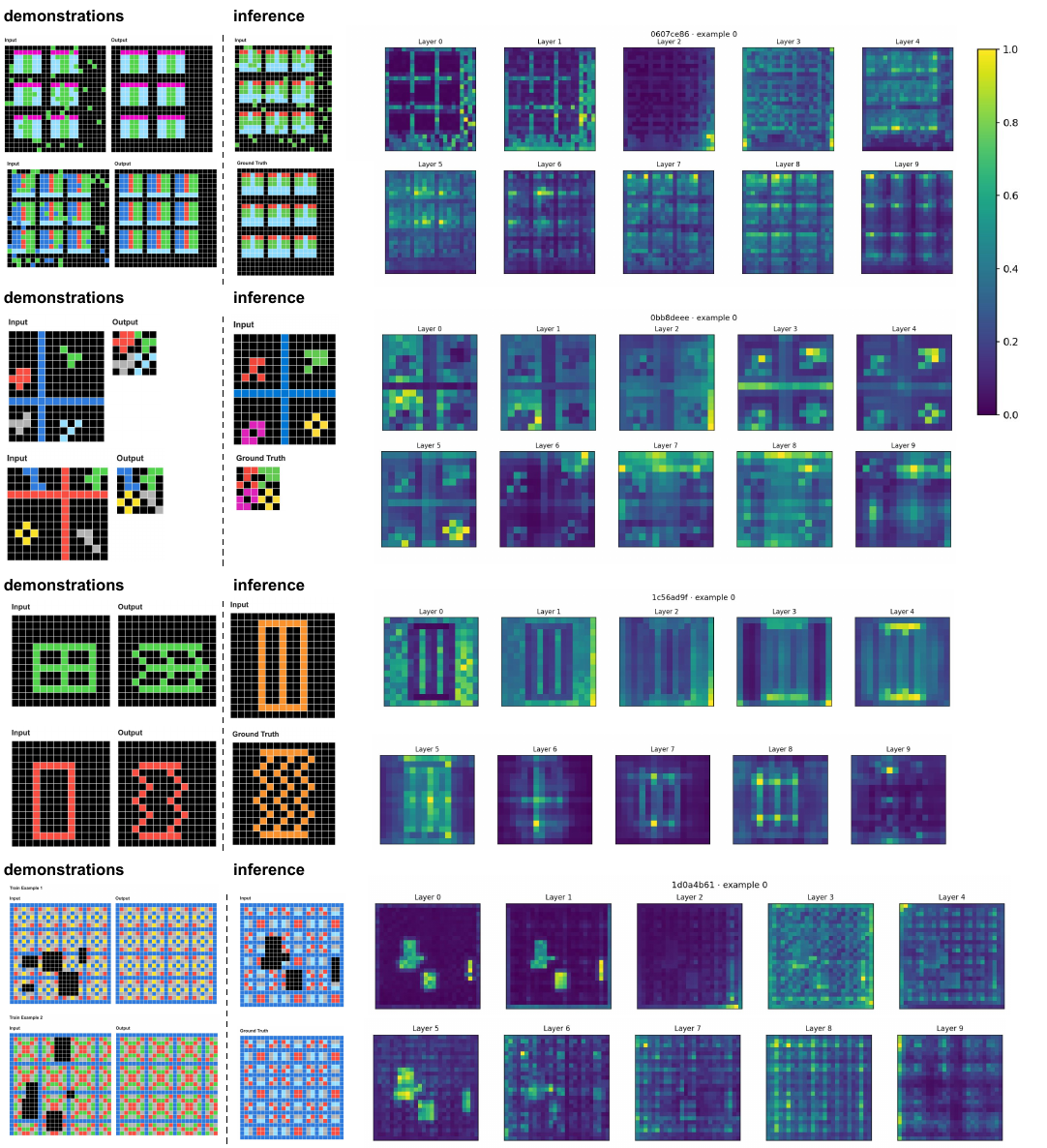}
  \caption{\textbf{Additional visualization: layer-wise attention maps.} Each map is the per-pixel softmax attention maps averaged across all pixels in that layer. 
  The corresponding demonstration examples (on the left) are provided for reference.
  }
  \label{fig:heatmap2}
\end{figure*}

\paragraph{Layer-wise Attention Maps.}
In~\cref{fig:heatmap2}, we visualize the layer-wise attention maps averaged across all pixels. See captions for detailed descriptions.

\subsection{Test-time Training Visualization}
\begin{figure*}
    \centering    \includegraphics[width=0.65\linewidth]{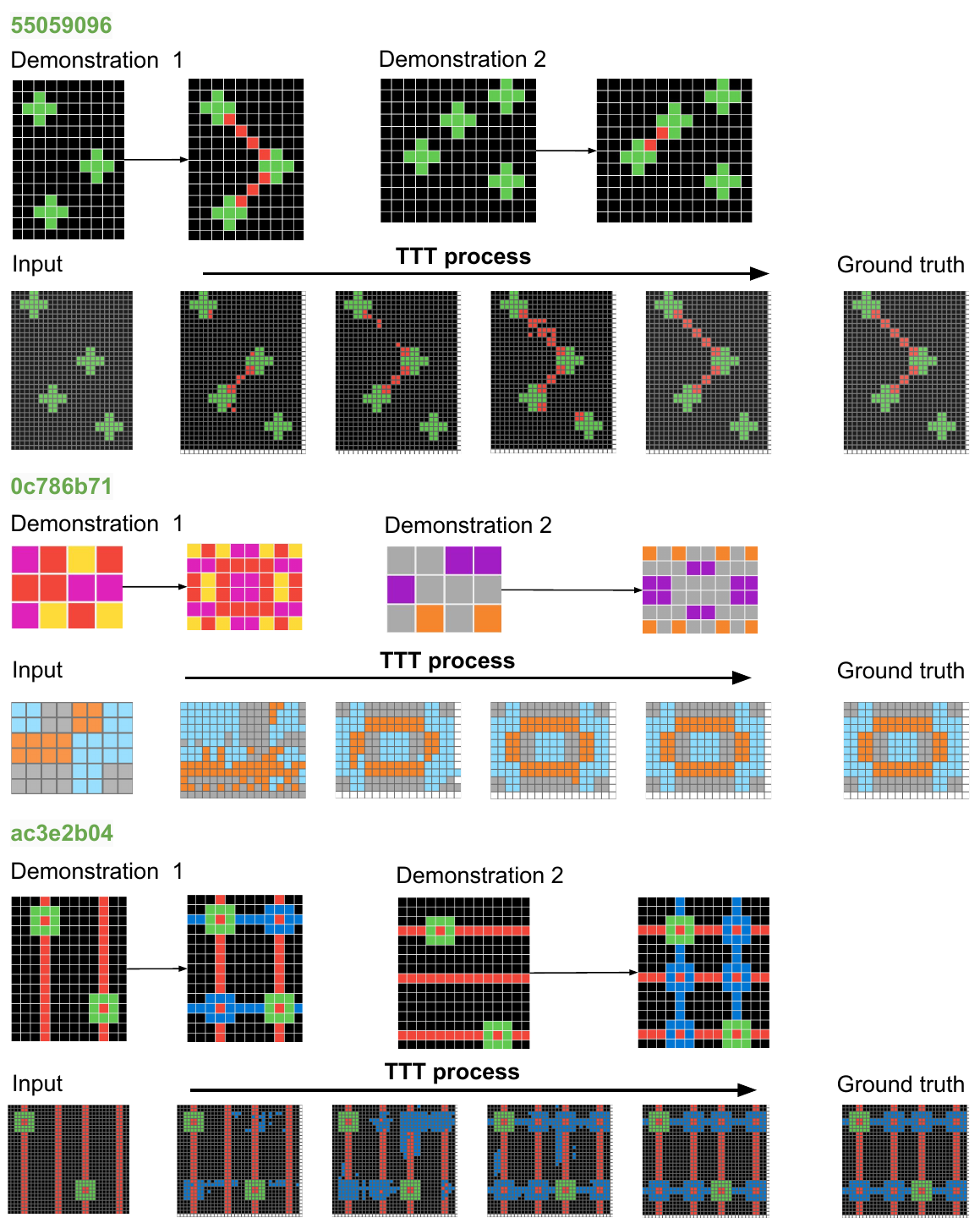}
    \caption{\textbf{Visualization of the test-time training process.} 
    Here, we visualize the grid augmented with a given scale ratio of 2$\times$ (the full canvas is not shown for brevity).
    As the test-time training progresses, the model’s predictions gradually converge toward the correct output. In early epochs, the model produces coarse and imprecise structures; in later epochs, the model can improve the solutions, \eg, by refining color and spatial arrangement. This visualization illustrates the model’s behavior of adapting to task-specific transformations through few-shot test-time training.
    }
    \label{fig:training_process}
\end{figure*}

Figure~\ref{fig:training_process} illustrates the evolution of model predictions during the test-time training process. Each row corresponds to a distinct test task from the ARC benchmark. It shows how our method progressively refines its prediction through test-time training.


\end{document}